\RequirePackage{fix-cm}
\documentclass[twocolumn]{svjour3}          
\smartqed  
\usepackage{graphicx}

\usepackage{subfigure}
\usepackage{amssymb}
\usepackage{lineno}
\usepackage{ulem}
\usepackage{indentfirst}
\usepackage{epstopdf}
\usepackage{color}
\usepackage{multirow}
\usepackage{mathptmx}
\usepackage{bm}
\usepackage[english]{babel}
\usepackage{blindtext}
\usepackage{mathrsfs}
\usepackage{amsmath,bm}
\begin{document}

\title{Robust Scene Text Recognition Using Sparse Coding based Features
}
\subtitle{}


\author{Da-Han Wang         \and 
Hanzi Wang \and
Dong Zhang \and
Jonathan Li \and
David Zhang
}


\institute{Da-Han Wang, Hanzi Wang, Dong Zhang \at
              School of Information Science and Engineering, Xiamen University, Fujian, China \\
              \email{dhwang@xmu.edu.cn, hanzi.wang@xmu.edu.cn}           
           \and
           Jonathan Li \at
              School of Information Science and Engineering, Xiamen University, Fujian, China \\
              Department of Geography and Environmental Management, University of Waterloo, Canada \\
              \email{ junli@uwaterloo.ca}
           \and
           David Zhang \at
           Department of Computing,The Hong Kong Polytechnic University, Hong Kong \\
           \email{csdzhang@comp.polyu.edu.hk}
}

\date{Received: date / Accepted: date}

\maketitle

\begin{abstract}
In this paper, we propose an effective scene text recognition method using sparse coding based features, called Histograms of Sparse Codes (HSC) features. For character detection, we use the HSC features instead of using the Histograms of Oriented Gradients (HOG) features. The HSC features are extracted by computing sparse codes with dictionaries that are learned from data using K-SVD, and aggregating per-pixel sparse codes to form local histograms. For word recognition, we integrate multiple cues including character detection scores and geometric contexts in an objective function. The final recognition results are obtained by searching for the words which correspond to the maximum value of the objective function. The parameters in the objective function are learned using the Minimum Classification Error (MCE) training method. Experiments on several challenging datasets demonstrate that the proposed HSC-based scene text recognition method outperforms HOG-based methods significantly and outperforms most state-of-the-art methods.
\end{abstract}

\section{Introduction}
\label{introduction}
\begin{figure*}[htbp]
\centering
\includegraphics[width=0.8\textwidth]{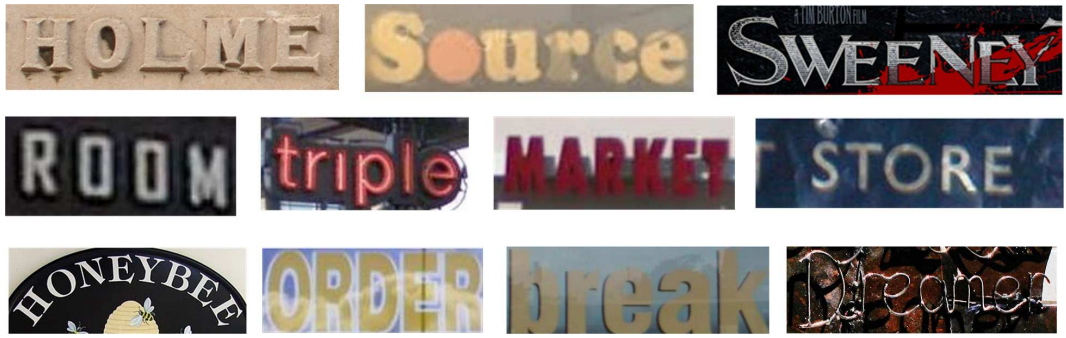}
\caption{ Examples of challenges in some scene text images. }
\label{examples1}
\end{figure*}

Since texts in images and videos contain rich high-level semantic information, which is valuable for understanding scenes, scene text recognition has attracted increasing attentions in the computer vision community in recent years. However, scene text recognition is nontrivial due to challenges such as background clutters, illumination changes, rotated characters, curved text lines, low resolution of characters, blurred characters, etc. Figure \ref{examples1} shows examples of some scene text images, showing the challenges of scene text recognition.

In scene text recognition, there are four main problems: (1) text detection, (2) text recognition, (3) full image word recognition, and (4) isolated scene character recognition. Text detection is to locate text regions in an image; while text recognition, given text regions, is usually referred to as cropped word recognition. Full image word recognition usually includes both text detection and text recognition in an end-to-end scene text recognition system. Isolated character recognition is usually a basic component of a scene text recognition system. In this paper, we mainly focus on the text recognition (or cropped word recognition) problem.

Since the work of \cite{WangK2011}, object recognition/detection based scene text recognition methods have achieved significant progress for scene text recognition \cite{Mishra2012a}\cite{WangT2012}\cite{Shi2013}\cite{Shi2014a}\cite{Shi2014b}. In this type of methods, each character class is considered as a visual object, and a character detector is used to detect and recognize character candidates simultaneously. The final word recognition result is obtained by combining character detection result and other contexts (e.g. geometric constraints, language model, etc.). Since this kind of methods jointly optimizes character detection and recognition, they have shown superior performance to traditional Optical Character Recognition (OCR) based methods.

In object recognition/detection based scene text recognition methods, a key issue is to design effective character features for character detection and classification. Since the Histograms of Oriented Gradients (HOG) features have been proven effective and popularly used in object detection \cite{Dalal2005}\cite{Felzenszwalb2010}, the HOG features have been introduced to character feature extraction for scene text recognition \cite{WangK2011}\cite{Mishra2012a}\cite{Mishra2013}. The core of HOG is to use gradient orientation at every pixel to represent the local appearance and shape of an object. While HOG is very effective in describing local features (such as edges) and robust to appearance and illumination changes, it can not effectively describe local structure information \cite{Zhang2011}.

In this paper, we mainly focus on the cropped word recognition problem, and propose an effective scene text recognition method using sparse coding based features, i.e., the Histograms of Sparse Codes (HSC) features that are originally proposed for object detection \cite{Ren2013}. In extracting the HSC features, per-pixel sparse codes  firstly computed with a dictionary, which is learned from data using K-SVD in an unsupervised way. The per-pixel sparse codes are then aggregated to form local histograms (similar to the HOG features). By representing common structures using the dictionary, the HSC features for character feature description have advantages that they can represent richer structural information than the HOG features.

For the character detector, we use simple classifiers including random ferns (FERNS), support vector machine (SVM), and sparse coding (SC) based classifier \cite{Wright2009}. The former two classifiers have been used for scene text recognition  \cite{WangK2011} \cite{Mishra2012a}, while the third one has not yet been used for scene text recognition. For word recognition, we design an objective function based on the Pictorial Structures (PS) model \cite{Felzenszwalb2005}\cite{Felzenszwalb2010} to integrate character detection result and geometric constraints. The parameters in the objective function are learned using the Minimum Classification Error (MCE) training method \cite{Juang1997}. Experiments on the popular ICDAR2003, ICDAR2011, SVT and III5K-Word datasets show that, for scene text recognition, the proposed method using the HSC features can significantly improve the performance of that using the HOG features, and it outperforms most of the state-of-the-art methods.

The main contributions of this paper are three-fold.
\begin{itemize}
\item[1.] First, we propose an effective scene text recognition method using sparse coding based features (i.e., the HSC features) that are learned automatically from data for character feature representation. Since the HSC features can represent richer structural information of character features, they can achieve superior performance in character/word recognition compared to the commonly used HOG features, as illustrated in our experiments (see Section \ref{experiments}). Since the HSC feature extraction method is as simple as HOG (see Section \ref{secHSC}), it can be considered as an alternative feature extraction method in applications (including character/word recognition) that need recognition, verification (detection), or classification.
\item[2.] Second, we show that the SC based classifier can achieve high performance in charater/word recognition, which reveals the potential of the SC based classifier in character/word recognition.
\item[3.] Third, we propose to use the MCE training method, which has been commonly used in speech recognition and handwriting recognition \cite{Juang1997} \cite{Biem2006}, to learn the parameters in the scene text recognition model. This provides a new way to learn parameters automatically in scene text recognition.
\end{itemize}

The rest of this paper is organized as follows. Section \ref{relatedWork} reviews the related works. Section \ref{secHSC} describes character feature representation method using sparse coding based features. The word recognition method is provided in Section \ref{wordRecModel}. Section \ref{experiments} presents the experimental results, and Section \ref{conclusion} offers concluding remarks. This paper is an extension of our previous work \cite{Zhang2014}. The extension includes that we provide more details and discussions about the proposed method, evaluate on more datasets and more aspects (e.g., the recognition speed, etc.), present more results on scene character recognition, and so on.

\section{Related Work} \label{relatedWork}

In scene text recognition, there are two main issues: one is to design a character detector that mainly involves feature representation and classification (whereas, much more attentions are paid to feature representation which will be reviewed in this section); the other one is to develop a word recognition model for integrating character detection and other contexts (i.e., word recognition or word formation). For classification, classifiers such as FERNS \cite{WangK2011}, SVM \cite{Mishra2012a}, random forest \cite{Yao2014} and CNN \cite{WangT2012}\cite{Jaderberg2014} have been adopted. In this section, we briefly review character feature representation methods and the main word recognition models, focusing more on the character feature representation methods.

\subsection{Character Feature Representation}
Feature representation is an important issue in pattern recognition and computer vision. For scene text recognition, feature representation attracts increasing attentions. Since scene character recognition is usually a basic component of a scene text recognition system, in this section we also review the main character feature representation methods proposed for scene character recognition. Existing character feature representation methods mainly fall in three types: HOG and its variants, mid-level character feature representation, and deep learning based methods.

The HOG features have been shown to be effective and have been used in object detection \cite{Dalal2005}, and for scene character feature representation \cite{WangK2011}\cite{Mishra2012a}\cite{Mishra2013}. Although HOG is very simple and effective in describing local features (such as edges), HOG ignores the spatial and structural information. Hence some methods are proposed to improve HOG. For example, Yi et al. \cite{Yi2013b} improve the HOG features by global sampling (called GHOG) or local sampling (called LHOG) to better model character structures.  Tian et al. \cite{Tian2013} propose the Co-occurrence of Histogram of Oriented Gradients (called CoHOG) features, which capture the spatial distribution of neighboring orientation pairs instead of only a single gradient orientation, for scene text recognition. The CoHOG method improves HOG significantly in scene character recognition.  Later, the authors of \cite{Tian2013} propose the pyramid of HOG (called PHOG) \cite{Tan2014} to encode the relative spatial layout of the character parts, and propose the convolutional CoHOG (called ConvCoHOG) \cite{Su2014} to extract richer character features by exhaustively exploring every image patches within a character image. These methods effectively improve the performance of scene character recognition.

Recently, some works propose to extract mid-level features for character feature representation. Yao et al. \cite{Yao2014} propose to use a set of mid-level detectable primitives (called strokelets), which capture substructures of characters, for character representation. The strokelets are used in conjunction with the HOG features for character description, as supplementary features to the HOG features. However, using the strokelets alone does not perform well. In \cite{Lee2014}, a discriminative feature pooling method that automatically learns the most informative sub-regions of each scene character is proposed for character feature representation. Gao et al. \cite{Gao2014a} propose a stroke bank based character representation method. The basic idea is to design a stroke detector for scene character recognition using the stroke bank. In \cite{Gao2014b}, Gao et al. propose to learn co-occurrence of local strokes by using a spatiality embedded dictionary, which is used to introduce more precise spatial information for character recognition. The results demonstrate the effectiveness of the two methods. All these methods explore mid-level features (such as the strokelets proposed in \cite{Yao2014}, the sub-regions proposed in \cite{Lee2014}, the stroke bank proposed in \cite{Gao2014a}, etc.) to represent character features, and have shown their effectiveness in scene characeter/text recognition.

The deep learning methods have also been adopted for feature learning of scene characters. Coates et al. \cite{Coates2011} propose a unsupervised feature learning method using convolutional neural networks (CNN) for scene character recognition. In \cite{WangT2012}, character features are extracted using an unsupervised feature learning algorithm similar to \cite{Coates2011}, and are integrated into a convolutional neural network (CNN) for character detection and classification. Recently, Jaderberg et al. \cite{Jaderberg2014} develop a CNN classifier that can be used for both text detection and recognition. The CNN classifier has a novel architecture that enables efficient feature sharing using a number of layers in common for character recognition. The performance achieved by deep learning based methods is pretty high on scene character and text recognition, showing the potential advantages of the deep learning based methods in scene character and text recognition. However, there still lacks theoretical analysis about the effectiveness of the CNN based methods, which needs further efforts of the community.

Character structure information is important to character representation. Besides Yao et al. \cite{Yao2014} and Lee et al. \cite{Lee2014} that exploit character structure features, Shi et al. \cite{Shi2013}\cite{Shi2014a}\cite{Shi2014b} propose to use part-based tree-structured features for representing character features. The part-based tree-structured features are designed directly according to the shape and structure of each character class. However, in \cite{Shi2013}\cite{Shi2014a}\cite{Shi2014b}, one needs to artificially design a tree-structured model and manually annotate training samples for each class. This is nontrivial and it is difficult to apply part-based tree-structured features to tasks with more character classes (such as Chinese characters).

The proposed HSC feature representation for scene text recognition is based on sparse coding, and is learned automatically, which can be viewed as a feature learning method using sparse coding. Compared to the feature learning method based on deep learning (such as CNN), the HSC feature extraction method is much more simpler and easier to implement.

In the context of feature learning, sparse coding has been a popular technique for learning feature representation, in applications such as image classification \cite{Yang2009}, object detection \cite{Kavukcuoglu2010}, etc. In \cite{Ren2013}, the HSC feature extraction method based on sparse coding is proposed for object detection showing its superior performance to the HOG features. However, it has not been applied to scene text recognition previously. To the best of our knowledge, this work is the first time to apply the HSC features to scene text recognition. We have surprisingly found that the use of the HSC features instead of the HOG features can significantly improve the performance of scene text recognition.

\subsection{Word Recognition Model}
Regarding the word recognition model for yielding word recognition results, Wang et al. \cite{WangK2011} apply a lexicon-driven pictorial structures model to combine character detection scores and geometric constraints. Mishra et al. \cite{Mishra2012a}\cite{Mishra2012b} build a conditional random field (CRF) model to integrate bottom-up cues (character detection scores) with top-down cues (lexicon prior). Similarly, Shi et al. \cite{Shi2013} use a CRF model to get final word recognition results. In \cite{Novikova2012}, the weighted finite-state transducers (WFSTs) based maximum a posteriori (MAP) inference is used for word recognition. In \cite{Goel2013}, the text is recognized by matching the scene and synthetic image features with a weighted dynamic time warping (wDTW) approach. In \cite{WangT2012}\cite{Phan2013}\cite{Neumann2013}\cite{Jaderberg2014}, heuristic integration models (i.e., summation of character detection scores) are used to integrate character detection result. In \cite{Shi2014b}, a probabilistic model is proposed to combine the character detection scores and a language model from the Bayesian decision view. In \cite{Weinman2014}, a semi-Markov model is used to integrate multiple information for scene text recognition. In \cite{Su2014b}, a word image is first converted into a sequential column vectors based on the HOG features, and then Recurrent Neural Network (RNN) is adapted to classify the sequential feature vectors into the corresponding word. The characteristics of the RNN based method proposed in \cite{Su2014b} is that it is able to recognize the whole word images without character-level segmentation and recognition.

It is also noticed that Almaz\'{a}n et al. \cite{Almazan2014} and Rodr\'{i}guez-Serrano et al. \cite{Serrano2015} propose to embed word attributes/labels and word images into a common subspace for word spotting and text recognition. Then the text recognition problem turns into a retrieval problem which essentially is a nearest neighbor search problem.  This idea provides a new solution to text recognition, and has show its effectiveness and efficiency \cite{Serrano2015}.

In this paper, we mainly focus on the character feature representation issue. For word recognition/formation, we simply apply the lexicon-driven pictorial structures model similar to that in \cite{WangK2011}. However, we improve it by taking into account the influence of the word length (i.e., the number of characters in the word) on word recognition results. Moreover, we propose to automatically learn the parameters of the word recognition model using the MCE training method to optimize the word recognition performance.

\section{Character Feature Representation based on Sparse Coding}\label{secHSC}

In this paper, we propose to learn a feature representation that describes richer structures of characters based on sparse coding. In \cite{Ren2013}, sparse coding is used to compute per-pixel sparse codes, which are aggregated into ``histograms'', i.e., forming the Histograms of Sparse Codes (HSC) features. HSC shows its advantages over HOG in that it can represent richer structures. We propose to use the HSC features to represent structures of characters for scene text recognition.

\subsection{Local Representation via Sparse Coding}

For computing per-pixel sparse codes, we need to learn a dictionary from data. We employ K-SVD \cite{Aharon2006} for dictionary learning, which learns common structures of objects in an unsupervised way. K-SVD is a generalization of the K-means algorithm and has been popularly used for dictionary learning in tasks such as image processing \cite{Elad2006}\cite{Romano2014}, image classification \cite{Yang2009}, face recognition \cite{Zhang2010}, action recognition \cite{Zheng2013}, handwritten digit recognition \cite{Zhang2013}, object recognition \cite{Jiang2013}, etc. In the following, we briefly introduce how to use K-SVD  to learn the dictionary for character feature representation.

In sparse representation, an input pattern is represented by a linear combination of atoms of a dictionary, under the constraints that the linear coefficients are sparse, which means that only a small fraction of entries in the coefficients are nonzeros. Denote a dictionary by $\mathbf{D}=[ \mathbf{d}_1, \  \mathbf{d}_2, \ ...\ ,\ \mathbf{d}_k ]$, where $\mathbf{d}_i \in \mathbb{R}^m (i=1,...,k)$ is the $i$-th basis vector and $k$ is the size of the dictionary. When a dictionary $\bf{D}$ is given, the coefficients for representing a given pattern ${\mathbf{x}}$ are computed by sparse coding, which can be written as:
\begin{equation} \label{sparseCoding}
\mathop {\min } \limits_{\alpha} \| \mathbf{x}-\mathbf{D}{\alpha} \|_2^2, \mbox{~subject~to~} ||\alpha||_0 \le T_0,
\end{equation}
where $\alpha$ is a sparse vector of the representation coefficients, and $T_0$ is a predetermined number of the non-zero entries in the coefficients for constraining the sparsity level. The obtained representation coefficients are referred to as the sparse codes of the pattern.

Besides sparse coding, another important issue in sparse representation is to learn the dictionary based on a set of training samples. Given a training data of $n$ samples, $\mathbf{X}={[{\mathbf{x}}_1, \ \mathbf{x}_2, \ ...\ ,\ \mathbf{x}_n]}, \ (\mathbf{X}\in \mathbb{R}^{m\times n})$, the dictionary learning problem can be addressed as \cite{Aharon2006}:
\begin{equation} \label{dictionaryLearn}
\mathop {\min }\limits_{\mathbf{D},\bm{\alpha} } \sum_{i=1}^n \frac{1}{2}\| \mathbf{x}_i-\mathbf{D}\alpha_i\|_F^2, \ \mbox{ subject to } \forall i, \ ||\alpha_i||_0<T_0, \\
\end{equation}
where $\| \cdot \|_F^2$ is the Frobenius norm defined as the sum of squares of the elements of the matrix/vector, and $\bm{\alpha}=[ {\alpha}_1, \ ...\ , \ {\alpha}_n ]$ is a $k\times n$ matrix. The problem can be rewritten as a matrix factorization problem with a sparsity constraint:
\begin{equation}
\mathop {\min }\limits_{\mathbf{D},\bm{\alpha} } \| \mathbf{X}-\mathbf{D}\bm{\alpha}\|_F^2, \ \mbox{ subject to } \forall i, \ ||\alpha_i||_0<T_0, \\
\end{equation}
 This is to search the best possible dictionary for the sparse representation of the sample set $\mathbf{X}$, and it is a joint optimization problem with respect to the dictionary $\mathbf{D}$ and the coefficients $\bm{\alpha}$. It can be solved by alternating between sparse coding of the samples based on the current dictionary and an update process of the dictionary elements.

The connection between sparse representation and clustering (i.e., vector quantization (VQ)) has been mentioned in previous works \cite{Delgado2003}\cite{Aharon2006}. In clustering, the representative examples (or the centers of clusters) can be viewed as the codewords of a dictionary (also called a codebook in vector quantization), and are used to represent samples using the nearest neighbor assignment. That is, when a dictionary is given, each sample is represented as its closest codeword. This can be considered as a special case of sparse coding, in the sense that only one atom participates in the reconstruction of the sample. Using the above notations, the sparse coding problem can be written as:
\begin{equation} \label{k-sc}
 {\min_{{\alpha}}} \|\mathbf{ x}- \mathbf{D} \alpha \|_2^2, \mbox{ subject to } {\alpha} \in \{ \mathbf{e}_1, \mathbf{e}_2, ..., \mathbf{e}_k \},
\end{equation}
where $\{ \mathbf{e}_1, \mathbf{e}_2, ...,\mathbf{ e}_k \}$ is the standard basis and $\mathbf{e}_i$ is a vector from the standard basis set with all zero entries except for a one in the $i$-th position. Similary, the codebook in vector quantization can be learned by minimizing the mean square error (MSE) as follows:
\begin{equation} 
\min_{\mathbf{D}, \bm{\alpha}} \| \mathbf{X} - \mathbf{D} \bm{\alpha} \|_F^2,  \mbox{ subject to } \ \forall i, \ \mathbf{x}_i=\mathbf{e}_j, \ \mbox{ for some } j.
\end{equation} The K-means algorithm is one of the most popular methods for learning codewords, which applies an iterative procedure with two steps in each iteration: 1) given $\mathbf{D}$, assign each sample to its nearest cluster, i.e., sparse coding; and 2) given that assignment, update $\mathbf{D}$ to further minimize MSE. Since MSE monotonically decreases in each iteration, K-means can guarantee at least a local optimum solution.

In \cite{Aharon2006}, the sparse representation problem is viewed as a generalization of the VQ problem, and the K-SVD algorithm is proposed to learn the dictionary by generalizing the K-means algorithm. If setting $T_0$ in Eqn. (\ref{dictionaryLearn}) to 1, the sparse representation problem becomes the VQ problem, as a special case. In the sparse coding step of K-SVD, any pursuit algorithm (such as the orthogonal matching pursuit (OMP) algorithm \cite{Pati1993}, the basis pursuit (BP) algorithm \cite{Chen2001}, etc.) can be adopted. In the second step, K-SVD updates one column of $\mathbf{D}$ at each time while fixing the other columns unchanged. The updated column and the new representation coefficients are obtained by minimizing MSE using singular value decomposition (SVD). The characteristics of K-SVD are that, it updates the columns of $\mathbf{D}$ sequentially and updates the relevant coefficients simultaneously. Though some other learning approaches apply similar procedures (i.e., the two steps), K-SVD is distinctive in that it updates each column separately by which it achieves better efficiency.

K-SVD is fast and efficient, and performs well in practice \cite{Robinstein2010}. It has been widely used in applications such as image processing, face recognition, object recognition, etc. In this paper, we use K-SVD to learn a dictionary that is used to compute per-pixel sparse codes in extracting the HSC features. In the sparse coding step of learning the dictionary using K-SVD, the OMP algorithm is adopted. In the dictionary update step, SVD performs once for updating one atom, and the procedure repeats $k$  times for a dictionary with $k$ atoms.

\begin{figure*}[htbp]
\begin{center}
  \includegraphics[width=0.9\textwidth]{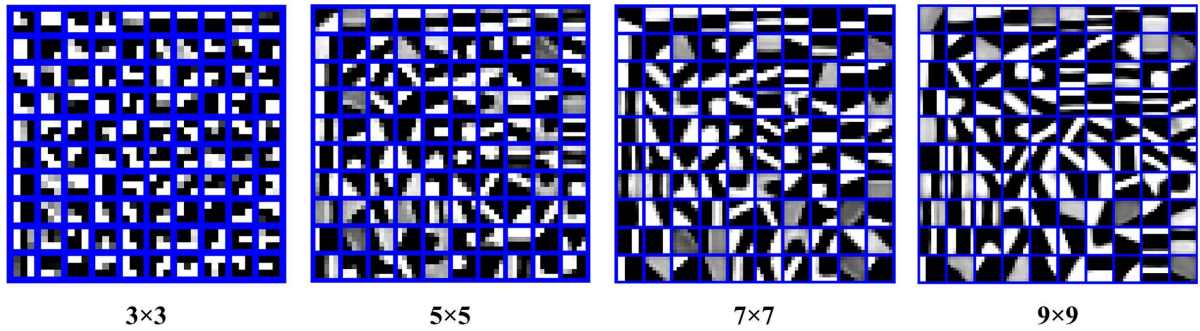}
  \caption{The dictionaries with different patch sizes learned by K-SVD.}
  \label{dic.eps}
\end{center}
\end{figure*}

For learning common structures, we use a set of image patches as the training set. The dictionary is learned by K-SVD. The Berkeley Segmentation Dataset and Benchmark (BSDS) \cite{Martin2001} is used to learn the dictionary. For obtaining the training image patches, we sample about 1000 image patches randomly from each image in BSDS. Then all the sampled image patches are used as the training set. Once the dictionary $\mathbf{D}$ is learned, sparse codes at each pixel in an image pyramid are computed using OMP. When learning the dictionaries, we can set the patch size and the dictionary size to different values (we will evaluate the influence of the dictionary size and the patch size in Section \ref{effectsSize}). In this paper, we set the dictionary size to $100$ and the patch size to $9 \times 9$. The learned dictionaries with four patch sizes are shown in Figure \ref{dic.eps}. We can see that, with a larger patch size, the dictionary can represent more rich and detailed structures.


\subsection{Aggregation into Histograms of Sparse Codes}
With learned sparse codes at each pixel, we aggregate them into histograms using a strategy similar to HOG. In extracting the HSC features, a character candidate is divided into small cells ($8\times 8$) and a feature vector of each cell is computed. Denote the sparse codes at a pixel by $\beta$. For each non-zero element $\beta_i$ in ${\beta}$, its absolute value $|\beta_i|$ is assigned to one of the four spatially-surrounding cells using bilinear interpolation. In each cell, a feature vector $F$ is obtained by averaging the codes in a $16 \times 16$ neighborhood. These features are called the HSC features in \cite{Ren2013}. $F$ is normalized with its $L_2$ norm. Finally, to increase the discriminative power of HSC, each value is transformed using the Box-Cox transformation \cite{Fukunaga1990} as follows:
\begin{equation}
\bar F = {F^\sigma }.
\end{equation}
We experimentally set the value of $\sigma$ to 0.25.

\begin{figure*}[htpb]
\centering
\subfigure[Character samples.]{
\begin{minipage}[t]{0.45\textwidth}
\includegraphics[width=1\textwidth]{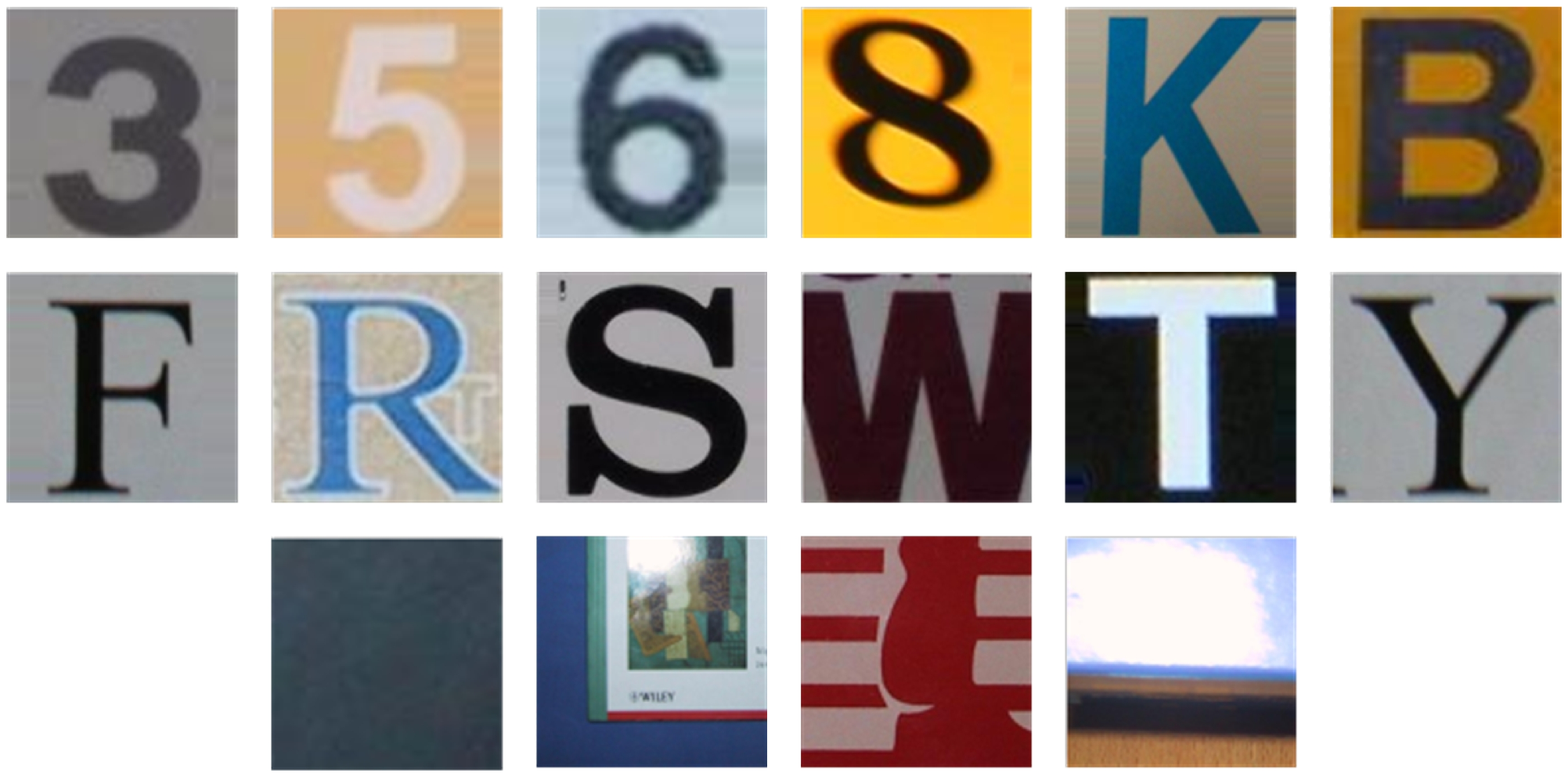}
\label{orig.eps}
\end{minipage}
}
\subfigure[HOG features of the character samples.]{
\begin{minipage}[t]{0.45\textwidth}
\includegraphics[width=1\textwidth]{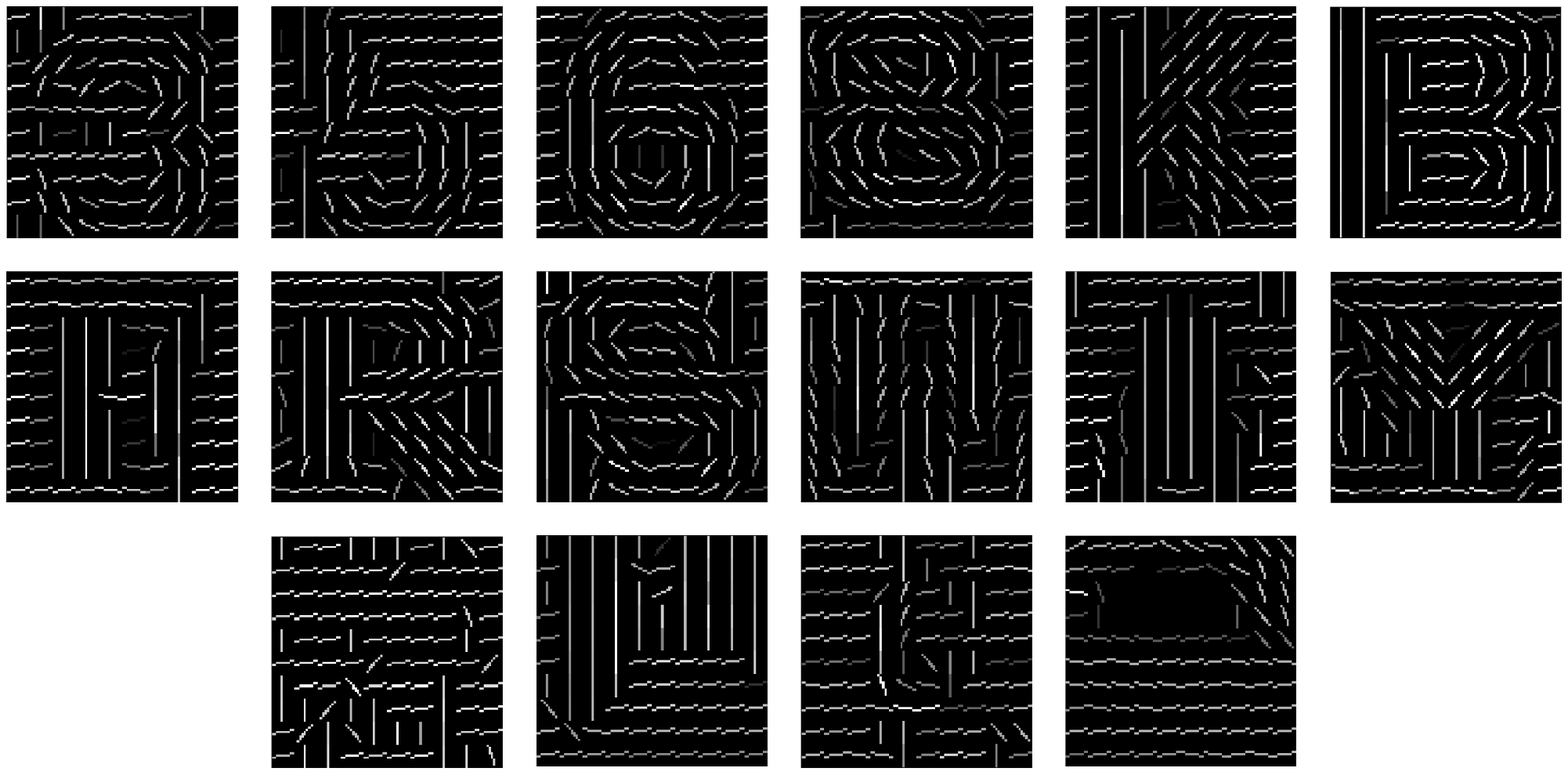}
\label{hog.eps}
\end{minipage}
}
\subfigure[HSC features of the character samples.]{
\begin{minipage}[t]{0.45\textwidth}
\includegraphics[width=1\textwidth]{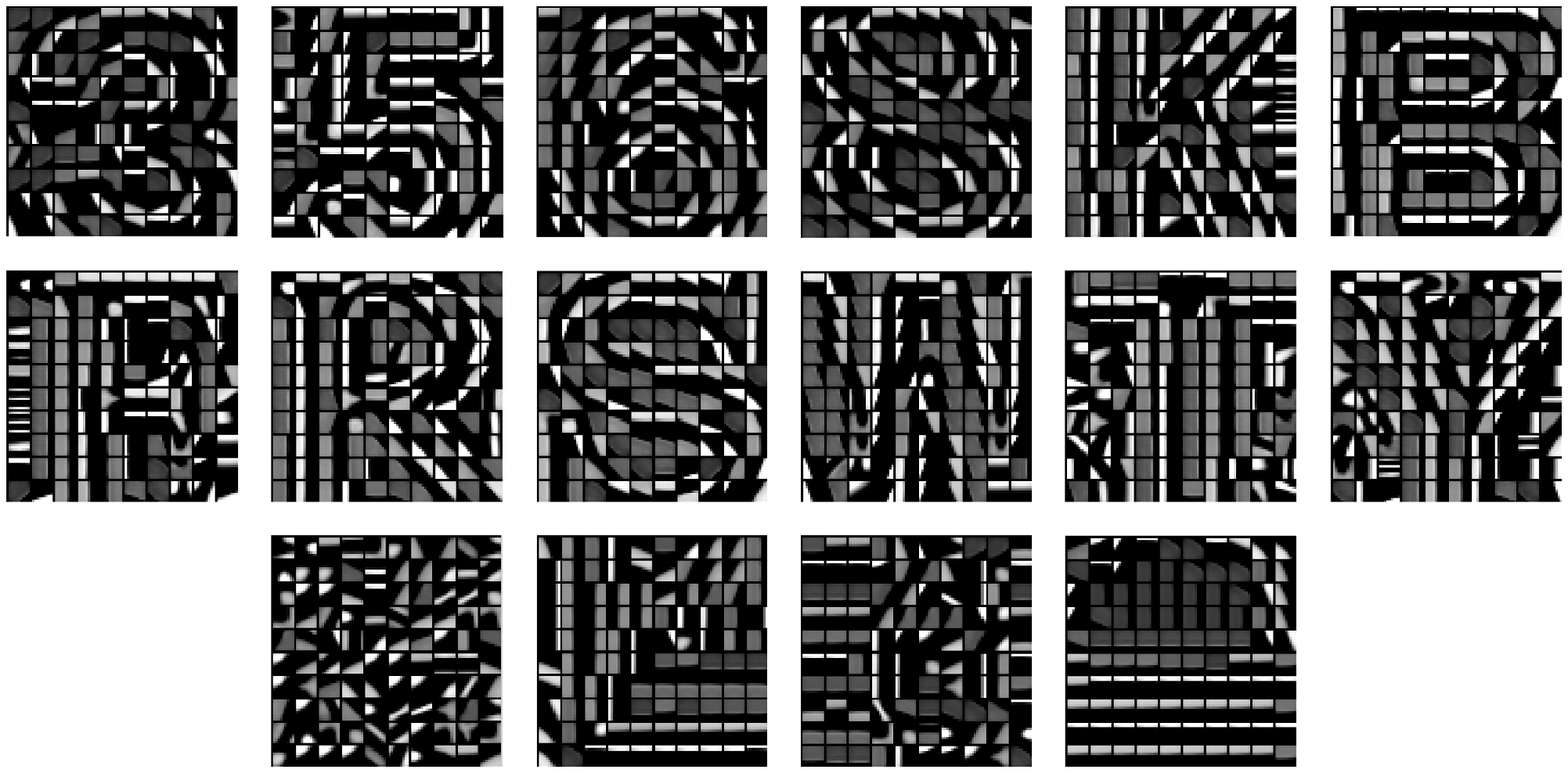}
\label{hsc.eps}
\end{minipage}
}
\caption{Visualizing the HOG and HSC features for some character and non-character samples: (a) Character samples (in the first two rows) and non-character samples (in the third row); (b) Dominant orientation in HOG, weighted by the gradient magnitude in the dominant orientation; (c) Dominant codeword in HSC, weighted by the histogram value of the dominant codeword. }
\label{visualize}
\end{figure*}

Figure \ref{visualize} shows the HSC and HOG features for some character samples as well as some non-character samples extracted from the background regions. From the figure, we can see that the HSC features capture richer structural information, and can better localize local patterns (such as textures, edges, corners, etc.) in each cell. In contrast, the HOG features capture less structural information and the edges in HOG may be off center.

\section{The Word Recognition Method} \label{wordRecModel}

\begin{figure*}[htbp]
\begin{center}
  \includegraphics[width=0.9\textwidth]{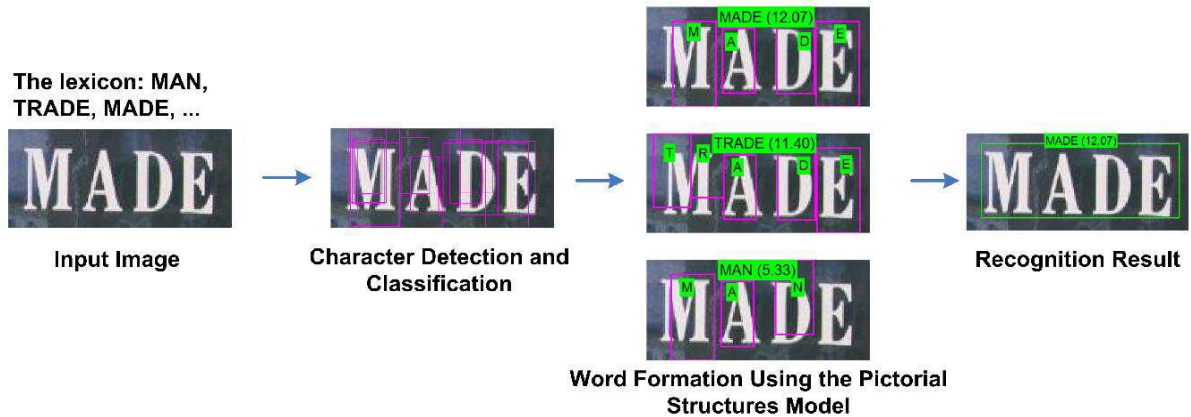}
  \caption{The detailed word recognition process using the word spotting strategy. The value in each candidate recognition result (e.g., 12.07 for ``MADE'', 11.40 for ``TRADE'', 5.33 for ``MAN'') denotes the matching score of the word to the image provided by the objective function, indicating the likelihood of the word.}
  \label{wordRec}
\end{center}
\end{figure*}

Figure \ref{wordRec} shows the main steps of the proposed method for word recognition. There are mainly two steps: character detection and classification, and word formation using the pictorial structures (PS) model. In the first step, for detecting and classifying character candidates, we train a character classifier based on the HSC features using a training set of character samples. In this step, we detect potential character candidates using a multi-scale sliding window strategy with the character classifier. Non-maximum suppress (NMS)\cite{Felzenszwalb2010} is adopted to get the final character detection result (see Section \ref{charDet} for more details).

After character detection and classification, character candidates are concatenated to form words. In this paper, we apply the lexicon-driven pictorial structures (PS) model \cite{Felzenszwalb2005} similar to that used in \cite{WangK2011} to form words. In object recognition using the PS model \cite{Felzenszwalb2005}, an object is represented by a collection of parts with connections between certain pairs of parts. For example, for faces, the parts are features such as the eyes, nose and mouth, and the connections can be described as the relative locations of these features. For people, the parts are the limbs, torso and head, and the connections allow for articulation at the joints. The PS model is suitable for the tasks of finding objects, where the best match of a PS model (related to the object) to an image is found. Hence, the PS model is an appealing model for word recognition, where characters can be viewed as the parts of a word and geometric relationships can be considered as the connections between the parts. In Figure \ref{wordRec}, the bounding boxes of characters of each candidate recognition result are provided. From the figure, we can see that each word can be considered as a collection of characters with each character being a part of the word.

In this paper, we use the word spotting strategy for word recognition. This means that for each word image, a lexicon consisting of a list of words is given, and word recognition is accomplished by finding the word that matches the image best. The matching score between a word and an image is given by an objective function that integrates character classification scores and geometric connections between characters based on the PS model. The match between a word and an image is searched using the Dynamic Programming (DP) algorithm based on the objective function. Section \ref{wordRecMod} gives more details. In Figure \ref{wordRec}, the matching results of three words (``MADE'', ``TRADE'', and ``MAN'') and the image are shown. The value in each candidate recognition result denotes the value of the objective function that evaluates the matching score between the input image and the given word. The word that yields the highest score is chosen as the recognition result.

\subsection{Character Detection and Classification} \label{charDet}
Character detection aims to detect character candidates using a character classifier. In this paper, we propose to apply the HSC features for character feature representation, and train the character classifier using the HSC features. We train a multi-class character classifier using a supervised training procedure. For each character class, the HSC features of the training samples are extracted using the learned dictionary. The HSC features of the whole training samples are fed into a classifier (such as the FERNS or SVM classifier) to train the character classifier.

The character detection algorithm is shown in Algorithm 1, which contains two steps: character candidate generation and classification, and non-maximum suppress (NMS) for obtaining the final character detection result. For each detected character candidate, the location of the character candidate, its corresponding character class and the classification score with respect to the character class are retained.

In the first step, the character candidates of each character class are detected separately. That is, for each character candidate, if its classification score with respect to a character class is larger than a threshold, it is considered as a character candidate of the character class. In our experiments, we set the threshold empirically to maintain moderate number of character candidates to guarantee that all true character regions are included while keeping the search space in the word formation step small (the more character candidates are maintained, the larger the search space is). In the second step, non-maximum suppress (NMS) is adopted for character detection. In this step, we use a simple greedy heuristic that is similar to \cite{Felzenszwalb2005}\cite{Felzenszwalb2010} to operate on all the character candidates in order: We first sort character candidates in descending order according to their classification scores; Then the NMS operation iterates on all the character candidates. If the character candidate has not yet been suppressed, we suppress all of its neighbors that are highly overlapped with the character candidate.

\linespread{1}
\small
\noindent\linethickness{2pt}
\line(1,0){240} \newline
\uline{\textbf{Algorithm 1. The Character Detection and Classification Algorithm}} \newline
\textbf{Input}: An image $I$, and a character classifier\newline
\textbf{Output}: Character detection result\newline
// \textbf{Step 1: Character Candidate Generation and Classification} \newline
Use the multi-scale sliding windows strategy to generate character candidates (the window size is set as 48 $\times$ 48, the step size for sliding is set as 8, and the input image is resized to a size that equals to the original size multiply the exponent of 0.5); \newline
For each character candidate $u_i$, $i=1,...,m$ ($m$ is the number of the generated character candidates)  \newline
\hspace*{0.5cm}Extract the HSC features of $u_i$; \newline
\hspace*{0.5cm}Perform character classification for $u_i$ using the proposed HSC-based character classifier, obtaining the output for \hspace*{0.5cm} each class: $O=\{O_w\}$ ($O_w$ denotes the output related to the character class $w$); \newline
\hspace*{0.5cm}For each character class $w$ \newline
\hspace*{1cm}If $O_{w}>thr$  \newline
\hspace*{1.5cm}Retain $u_i$, $w$, and $O_{w}$; \newline
\hspace*{1cm}End if \newline
\hspace*{0.5cm}End for \newline
End for \newline
//\textbf{Step 2: Non-Maximum Suppress (NMS) } \newline
Sort the retained character candidates in descending order according to their classification scores; \newline
For each character candidate $u_j$ with its score $O_{w_j}$, $j=1,...,n$ ($n$ is the number of the retained character candidates) \newline
\hspace*{0.5cm}If $u_j$ is not suppressed, \newline
\hspace*{1cm}Get the neighbors of $u_j$ according to the overlap degree; \newline
\hspace*{1cm}Suppress the neighbors of $u_j$; \newline
\hspace*{0.5cm}End if \newline
End for \newline
Output the final character detection result. \newline
\uline{\hspace*{8.5cm}}
\normalsize

The output of the character classifier is given as follows:
\begin{equation} \label{probRatio}
 S(w, u)=\log(\frac{p(w|u)}{p(w_{bg}|u)})=\log p(w|u)-\log p(w_{bg}|u),
\end{equation}
where $u$ denotes the character candidate, $w$ denotes the character class, and $w_{bg}$ denotes the background class. The authors of \cite{WangK2011} have shown the effectiveness of Eqn. (\ref{probRatio}) in object detection.

\subsection{The Word Recognition Model} \label{wordRecMod}
In the character detection and recognition step, character candidates of each character class are maintained. In the word formation step, character candidates are used to form the word. As indicated previously, the match of each word in the lexicon to the image is found using the DP algorithm. The procedures of concatenated character candidates corresponding to a given word are as follows. Let $W=(w_1, w_2,...,w_n)$ be a word with $n$ characters (note that $w_i (i=1,...,n)$ denotes a character class). Let $u_1$ be one of the character candidates of the character class $w_1$, $u_2$ be one of the character candidates of the character class $w_2$, and so forth. Then the character candidate sequence $u_1, u_2,..., u_n$ will be a detected word candidate of $W$, which is called a configuration of $W$ \cite{Felzenszwalb2005}\cite{WangK2011}. All the configurations will be evaluated by an objective function that integrates the character classification scores and geometric relationships between pairs of adjacent characters. The optimal configuration can be obtained using the DP algorithm based on the objective function. We also use simple rules to reduce the searching space, such as the horizontal/vertical distance between the character candidates of two successive characters (such as $u_i$ and $u_{i+1}$) should not be larger than three times the width of $u_i$ and the two successive characters $u_{i+1}$, $u_i$ and $u_{i+1}$ should not highly overlapped, etc.

For word recognition, we design a new objective function based on the PS model \cite{Felzenszwalb2010} to evaluate each word in the lexicon. The PS model has been used for scene text recognition in \cite{WangK2011} to find an optimal configuration of a given word. However, in \cite{WangK2011}, the objective function does not consider the word length, which leads to a drawback that words' scores are influenced by their lengths, and thus words of different lengths are not comparable. We improve it by considering the word length in the objective function. Since when using the word spotting strategy for word recognition, the list of words is provided (which means that the prior probabilities of the words in the lexicon are identical), we do not integrate an English language model in the objective function.

Let $S(w_i, {u}_i)$ be the classification score given by a character classifier as in Eqn. (\ref{probRatio}). The objective function is designed as follows:
\begin{equation} \label{objectFunc}
O= \sum\limits_{i = 1}^n { S(w_i, {u}_i)} + \lambda_1 \sum\limits_{i=1}^{n-1} {Z({u}_{i},{u}_{i+1})} + \lambda_2 n,
\end{equation}
where $Z({u}_{i},{u}_{i+1})$ is a geometric model to evaluate the compatibility of $u_i$ and $u_{i+1}$ in geometric constraints, and $\lambda_1(>0)$ and $\lambda_2(<0)$ are the coefficients to balance the contributions of $Z$ and the word length $n$ to the objective function. The last term in (\ref{objectFunc}) can be viewed as a penalty term, and is used to overcome the bias caused by long words. The geometric model $Z$ is modeled by a linear SVM classifier, and the extracted features for $Z$ include scale similarity, overlapping of the two candidates, distance between the top positions and the bottom positions of the two candidates, etc. The parameters (i.e., $\lambda_1$ and $\lambda_2$) in the objective function in Eqn. (\ref{objectFunc}) are learned using the Minimum Classification Error (MCE) training method that has been widely used in speech recognition \cite{Juang1997} and handwriting recognition \cite{Biem2006} \cite{Liu2004c} \cite{Wang2012a}, as introduced in the following section.

\subsubsection{Parameter Learning}
In word-level MCE training, the coefficients in Eqn. (\ref{objectFunc} ) are estimated using a training dataset which contains $R$ scene text images, denoted by $ D_I=\{ (I^s, W_t^s, U^{s}_t|s=1,...,R \}$, where $I^s$ is the word image, $W_t^s=(w_{t1}^s...w_{tn}^s)$ ($n$ is the number of characters in $W_t^s$) is the ground-truth transcript of $I^s$, and $U_t^{s}=(u_{t1}^{s}...u_{tn}^{s})$ is the ground-truth character candidate sequence of the character class sequence ($w_{t1}^s...w_{tn}^s$).

Following the MCE training procedure \cite{Juang1997}, the misclassification measure on a cropped word image sample is estimated by:
\begin{equation} \label{misMeasure}
d(I,\Lambda)=-g(U_t, W_t, \Lambda)+g(U_r,W_r,\Lambda),
\end{equation}
where $\Lambda$ is the parameter set, $g(U_t, W_t, \Lambda)$ is the discriminant function for the ground-truth word $W_t$ and the ground-truth configuration $U_t$, and $g(U_{r},W_r,\Lambda)$  is the discriminant function of the closest rival word $W_r$ and its optimal configuration $U_r$:
\begin{equation}
g(U_{r},W_r,\Lambda) = \mathop {\max }\limits_{{W_k\in L,(U_k,W_k)\ne(U_t,W_t)} } g(U_k,W_k,\Lambda ), \nonumber
\end{equation}
where $L$ is the provided lexicon for the image $I$.
The misclassification measure is then transformed by the sigmoidal function as:
\begin{equation} \label{lossFunction}
l(I,\Lambda ) = \frac{1}{{1 + {e^{ - \xi d(I,\Lambda )}}}},
\end{equation}
where $\xi$ is a parameter to control the hardness of sigmoidal nonlinearity and it is usually set to 1. The parameters in MCE training are learned using the stochastic gradient descent algorithm (SGD) \cite{Robbins1951} on each input sample.

For scene text recognition, the discriminant function $g$ is the designed objective function as Eqn. (\ref{objectFunc}). The rival word is the one that is the most confusable word with the correct word, and is obtained using the beam search algorithm \cite{Wang2012a}. The parameters are updated iteratively by SGD as follows:
\begin{equation} \label{updateFormula}
\begin{array}{l}
\Lambda (t + 1)\; = \Lambda (t) - \varepsilon (t){{\partial l(U,\Lambda )} \over {\partial \Lambda }}{|_{\Lambda  = \Lambda (t)}}\\
{\rm{~~~~~~~~~~~~~~}} = \Lambda (t) - \varepsilon (t)\xi l(1 - l){{\partial d(U,\Lambda )} \over {\partial \Lambda }}{|_{\Lambda  = \Lambda (t)}}\\
{\rm{~~~~~~~~~~~~~~}} = \Lambda (t) - \varepsilon (t)\xi l(1 - l)({O_r} - {O_t}),
\end{array}
\end{equation}
where $\varepsilon (t)$ is the learning rate.

\section{Experiments}\label{experiments}
We evaluate the proposed HSC-based scene text recognition method on several popular datasets including the ICDAR2003 \cite{Lucas2003}, ICDAR2011 \cite{Shahab2011}, Stree View Text (SVT) \cite{WangK2011}, and III5K-Word datasets \cite{Mishra2012b}.  The ICDAR2003 dataset contains 507 natural scene images (including 258 training images and 249 test images) in total. The images are annotated at character level. Characters and words can be cropped from the images. The ICDAR2011 dataset contains 229 images for training and 255 images for test. The SVT dataset is composed of 100 training images and 249 test images. For the ICDAR2011 and SVT datasets, only the words in the images can be cropped because the images are annotated at word level only. The IIIT5K-Word dataset is the largest and most challenging dataset for word recognition so far. This dataset includes 5000 word images, where 2000 images are used for training and 3000 images for test. For fair comparison with previous works, we ignore non-alphanumeric characters and words with 2 or fewer characters in word recognition when using the ICDAR2003, ICDAR2011, and SVT datasets. The details of these datasets are shown in Table \ref{datasets}.

Our experiments are implemented on a PC with an Intel(R) Core(TM) i7-2670QM CPU 2.20 GHz processor and 8 GB RAM, and are programmed using Matlab R2011a.

\linespread{1}
\begin{table*}[htbp]
\small
\caption{  Datasets used for evaluation. }
\label{datasets}
\begin{center}
\begin{tabular}{|c|cc|cc|cc|cc|}
\hline
   	 & \multicolumn{2}{c|}{SVT} & \multicolumn{2}{c|}{ICDAR03} &\multicolumn{2}{c|}{ICDAR11}  &\multicolumn{2}{c|}{III5K-Word} \\ \hline
   & training &test &training &test &training &test &training &test\\ \hline
   \#Images &100 &249 &258 &249 &229 &255 & -- &-- \\
      \#Words &258 &647 &1,156 &1,107 &846 &1,189 & 2000 &3000 \\
   \hline
\end{tabular}
\end{center}
\end{table*}

\subsection{Character Classification}\label{charClassification}
In this section, we first evaluate the performance of the proposed HSC-based character classifier in the task of character classification. 

\subsubsection{Classifier Training} \label{sec-clsfTrn}
For training the character classifier, we get the hybrid training set by combining the cropped characters from the training set of the ICDAR2003 dataset (6,185 samples), the training character samples of Chars74K-15 \footnote{http://www.ee.surrey.ac.uk/CVSSP/demos/chars74k/.}  (930 samples), and the training samples of the synthetic data (60,219 samples) produced by Wang.~et.~al \cite{WangK2011}. For better detection performance, we consider the background  images as one class (i.e., the background class or the non-character class), and a classifier with 63 classes is used for character detection. We use the background image regions that are extracted from the training images of the ICDAR2003 dataset as the training samples of the non-character class. To increase the size of the training samples of the non-character class, the images in the Microsoft Research Cambridge Object Recognition Image Database \footnote{http://research.microsoft.com/en-us/downloads/b94de342-60dc-45d0-830b-9f6eff91b301/default.aspx.} are added as the training samples of the non-character class as well. In evaluating the performance of the character classifiers, two datasets are used for test: ICDAR03-CH and Chars74K-15. The former dataset consists of the character samples cropped from the test set of the ICDAR2003 dataset, which totally contains  5,379 character samples. The latter one contains 930 test samples (15 samples per character class).

We propose to use the HSC features in the character classifier. To test the influence of different features on the performance of the proposed method, we also use the HOG features and the LBP features \cite{Ojala2002} in the character classifier for comparison.

For the HOG features, we use the Vlfeat library \cite{Vedaldi2008} for the HOG feature extraction. We use a variant of the HOG features proposed in \cite{Felzenszwalb2010}. To generate feature vectors with the same length, we first resize each training sample into the size of $48 \times 48$, then divide the image into square cells of size $8 \times 8$, obtaining $6 \times 6$ cells. In each cell, a 31-dimensional vector is extracted by aggregating per-pixel gradients (see \cite{Felzenszwalb2010} for more details). Finally, we obtain a feature vector with a dimensionality of 1116 for each sample.

For the HSC features, we also resize each training sample into the size of $48 \times 48$, and the size of each cell is set to $8\times 8$ as well. For each cell, a feature vector with a dimensionality that is the same as the dictionary size (i.e., $k$) is extracted. Since the feature vector in each cell is obtained by averaging the codes in a $16\times16$ neighborhood, the cells on the boundary are not used to form the feature vector of the whole image. Hence, the dimensionality of the extracted HSC features on each image sample is $4 \times 4 \times k$, which is 1600 by default.

For the LBP features, we adopt the implementation of Vlfeat as well. Each training sample is also resized into the size of $48 \times 48$, and dividied into square cells of size $8 \times 8$. In each cell, a 58-dimensional vector is extracted. The dimensionality of the extracted LBP features for each image sample is 2088.

To test the influence of classifiers on the performance of the proposed method, we also evaluate several classifiers. The classifiers we test in this paper include FERNS, linear SVM, and SC. The FERNS classifier is trained as in \cite{WangK2011}. In sparse coding, for each character class, we learn a dictionary using the training samples of the class. In classification, an input character pattern is classified into the class that gives the minimum reconstruction error, similar to \cite{Wright2009}. For implementation, we use the SPArse Modeling Software (SPAMS) developed by Mariral et al. \cite{Mairal2010} for dictionary learning and sparse coding, and classifiers with various number of basis vectors (including 10, 20, 30, 50, and 100, denoted by SC-10, SC-20, ..., SC-100, respectively) are evaluated.

We have conducted some experiments to evaluate the performance of the character classifier on classifying non-character regions. We first randomly sampled some non-character regions (with size 48*48) from the test dataset of the ICDAR2003 dataset, obtaining about 5,000 non-character samples (which is disjoint with the non-character samples for training). We test the performance using the HOG and HSC features and the FERNS, SVM, and SC classifiers. The results show that the performance is quite high for SVM and SC (higher than 96\%) while the FERNS classifier performs much worse (about 85\%). Comparing the HOG features and HSC features, we can find that the HSC features and the HOG features perform comparable. This shows that the classifier can reject most regions that are not like a character region. Due to the page limitation, we do not put these results in the paper.

\subsubsection{Classification Results}

The character classification results using different classifiers trained with the hybrid training set and different features are shown in Table \ref{charClsf}. As we can see, the HSC features outperform the HOG features and the LBP features significantly for all the classifiers on the two datasets. For the FERNS classifier, the increase of classification accuracy is about 14\% on ICDAR03-CH and is about 10\% on Chars74K-15 when the HSC features are used instead of the HOG features, and  the increase of recognition accuracy is about  37\% on ICDAR03-CH and is about 27\% on Chars74K-15 when the HSC features are used instead of the LBP features. For the other classifiers, the improvements of recognition accuracy using the HSC features are also obvious.

Comparing the performance of different classifiers, we can see that the SC classifiers with more than 30 basis vectors perform better than the FERNS classifier and the SVM classifier on the two datasets. When using the HSC features, the SC-50 classifier yields the best performance which is comparable with that obtained by the SC-100 classifier on the ICDAR03-CH dataset, while the SC-100 classifier performs the best on the Chars74K-15 dataset. When using the HOG features and the LBP features, the SC-100 classifier performs the best on both datasets. Though only using a linear kernel, the SVM classifier shows promising performance for all the three features.

The LBP features perform much worse than the HOG features and the HSC features. This is mainly due to the fact that the LBP features describe texture information, while for character classification, gradients and other structural information (such as edges, corners, etc.) play a more important role. Hence, the performance of the LBP features is poor, indicating that the LBP features are not suitable for character classification. Thus, in the following experiments, we do not consider the LBP features.

\linespread{1}
\begin{table*}[htbp]
\small
\caption{Results of character classification using different classifiers trained with the hybrid training set (\%).}
\label{charClsf}
\begin{center}
\begin{tabular}{|l|l|ccccccc|}
\hline
 &  & FERNS & SVM & SC-10 & SC-20 & SC-30 & SC-50 & SC-100 \\
\hline
\multirow{3}*{ICDAR03-CH} & HOG & 52.80 & 71.65 & 71.09 & 74.76 & 75.82 & 77.07 & 77.76 \\
                               & LBP & 29.04 & 65.92 & 62.30 & 65.22 & 66.03 & 65.80 & 66.60 \\
                                & HSC & \textbf{66.79} & \textbf{77.28} & \textbf{76.66} & \textbf{79.62} & \textbf{80.03} & \textbf{81.13} & \textbf{81.10} \\
\hline
\multirow{3}*{Chars74K-15} & HOG & 37.20 & 59.14 & 50.11 & 58.49 & 60.75 & 62.04 & 64.73 \\
                              & LBP & 18.49 & 49.14 & 41.61 & 45.48 & 45.16 & 45.27 & 45.48 \\
                              & HSC & \textbf{46.56} & \textbf{64.95} & \textbf{60.00} & \textbf{64.41} & \textbf{66.67} & \textbf{67.53} & \textbf{68.17} \\
\hline
\end{tabular}
\end{center}
\end{table*}

\subsubsection{Experiments with Individual Training Datasets}
For fair comparison with previous works, we also use individual training datasets to train the character classifier. That is, for the ICDAR2003 dataset, we use the training set of the ICDAR2003 dataset to train a classifier, and test the classification accuracy on the test set of the ICDAR2003 dataset. For the Chars74K-15 dataset, we use the training set (15 training samples per character class) of the Chars74K-15 dataset to train a classifier, and evaluate on the test set of the Chars74K-15 dataset. These training and test settings are consistent with those used in previous works \cite{Lee2014}\cite{Campos2009}\cite{Lucas2003}.

The classification results using different classifiers trained with individual training datasets and different features are shown in Table \ref{charClsf2}. The classification results in Table \ref{charClsf2} also show the superiority of the HSC features to the HOG features. The difference between the results in Table \ref{charClsf} and those in Table \ref{charClsf2} lies in the performance of the SC classifier. In Table \ref{charClsf}, the tendency of the performance of the SC classifier is that, the more the number of basis vectors is, the better its performance is. However, in Table \ref{charClsf2}, the tendency is not obvious, and the performance of SC-100 is even worse (on the Chars74K-15 dataset, the SC-100 classifier does not even work). This is  because for the Chars74K-15 dataset, only 15 training samples for each character class are used to train the classifier, which will cause the overfitting problem in training the SC classifier with 100 basis vectors. Due to the limited number of the training samples of the ICDAR2003 and Chars74K-15 datasets, the SC classifier does not benefit much from using a larger number of basis vectors. However, in training the character classifier with the hybrid training set, the number of the training samples for each class is much larger and thus the SC classifier with a larger number of basis vectors performs better.

\linespread{1}
\begin{table*}[htbp]
\small
\caption{Results of character classification using different classifiers trained with individual training datasets (\%).}
\label{charClsf2}
\begin{center}
\begin{tabular}{|l|l|ccccccc|}
\hline
 &  & FERNS & SVM & SC-10 & SC-20 & SC-30 & SC-50 & SC-100 \\
\hline
\multirow{2}*{ICDAR03-CH} & HOG & 40.85  & 72.92 & 71.09 & 74.78 & 75.73 & 75.86 & 73.12 \\
                               & HSC & \textbf{57.20} & \textbf{77.50} & \textbf{78.62} & \textbf{78.88} & \textbf{78.52} & \textbf{78.38} & \textbf{74.52} \\
\hline
\multirow{2}*{Chars74K-15} & HOG & 43.87 & 59.03 & 58.92 & 59.68 & 59.46 & 59.03 & -- \\
                              & HSC & \textbf{56.13} & \textbf{67.96} & \textbf{68.06} & \textbf{67.96} & \textbf{68.49} & \textbf{64.19} & -- \\
\hline
\end{tabular}
\end{center}
\end{table*}

In this paper, since we do not concern the influence of the number of the training samples on the performance of character classification and word recognition, we simply use the classifiers trained with the hybrid training set for the experiments of word recognition (see Section \ref{sub-wordrec}) because training with more samples generally provides better performance. The experiments with individual training datasets are only for fair comparison with previous works in the task of character classification to illustrate the effectiveness of the HSC features.

We compare the proposed method with several state-of-the-art methods on the two datasets, as shown in Table \ref{compCharClsf}. In the table, we report the classification accuracies of the HSC features and the SC/SVM classifier trained with the hybrid training set or individual training datasets. The compared methods include: the method (ConvCoHOG+Linear SVM) \cite{Su2014} that uses the ConvCoHOG features and a linear SVM classifier for classification, the method (PHOG+Linear/Chi-Square SVM) \cite{Tan2014} that uses the PHOG features and a SVM classifier  with the linear kernel or the chi-square kernel for classification, the method (CoHOG+Linear SVM) \cite{Tian2013} that uses the CoHOG features and a linear SVM classifier for classification, the method (HOG+AT+Linear SVM) \cite{Mishra2013} that uses the HOG features (applying affine transform (AT) to enrich the training samples) and uses a linear SVM classifier for classification, the method (Feature Pooling+L2 SVM) \cite{Lee2014} that uses a mid-level feature pooling method to learn informative sub-regions of characters and adopts a L2-regularized SVM classifier for classification, the methods (GHOG+Chi-Square SVM and LHOG+Chi-Square SVM) \cite{Yi2013b}, the methods (HOG+NN and HOG+FERNS) \cite{WangK2011} that use the HOG features and the nearest neighbor (NN) classifier or the FERNS classifier, the method (MKL) \cite{Campos2009} that uses the multiple kernel learning (MKL) approach to combine different features, the method (GB+RBF SVM) \cite{Campos2009} that adopts the Geometric Blur (GB) feature \cite{Berg2005} and the SVM classifier with a RBF kernel, and the method (ABBYY) \cite{Campos2009} that uses the commercial OCR system ABBYY FineReader \footnote{http://www.abbyy.com} for classification. In the table, we mark the methods that use individual training datasets to train a classifier with a asterisk (*) for convenience, while we do not mark the methods that use extra training datasets.

\linespread{1}
\begin{table*}[htbp]
\small
\caption{Comparing the proposed method with several state-of-the-art methods on character classification (\%).}
\label{compCharClsf}
\begin{center}
\begin{tabular}{|l|cc|}
\hline
Method & Chars74K-15 & ICDAR03-CH \\
\hline
Proposed HSC+SC (using the hybrid training set) & \textbf{68} & \textbf{81} \\
Proposed HSC+Linear SVM (using the hybrid training set) & 65 & 77 \\
Proposed HSC+SC  (*) & \textbf{68} & 79 \\
Proposed HSC+Linear SVM (*) & \textbf{68} & 78 \\
\hline
Deep CNN \cite{Jaderberg2014} & \textbf{80.3} & \textbf{91.0}  \\
ConvCoHOG+Linear SVM \cite{Su2014} & - & \textbf{81} \\
CoHOG+Linear SVM \cite{Tian2013} & - & 79.4 \\
PHOG+Chi-Square SVM \cite{Tan2014} & - & 79 \\
PHOG+Linear SVM \cite{Tan2014} & - & 76.5 \\
HOG+AT+Linear SVM \cite{Mishra2013}  & \textbf{68} & 73 \\
Feature Pooling+L2 SVM \cite{Lee2014} (*) & 64 & 79 \\
GHOG+Chi-Square SVM \cite{Yi2013b} (*) & 62 & 76 \\
LHOG+Chi-Square SVM \cite{Yi2013b} (*) & 58 & 75 \\
HOG+NN \cite{WangK2011} (*) & 58 & 52 \\
MKL \cite{Campos2009} (*) & 55 & - \\
HOG+FERNS \cite{WangK2011} (*) & 54 & 64 \\
GB+RBF SVM \cite{Campos2009} (*) & 53 & - \\
ABBYY \cite{Campos2009} (*) & 31 & 21 \\
\hline
\end{tabular}
\end{center}
\end{table*}

Comparing the competing methods using individual training datasets, we can see that the proposed method achieves promising performance on the two datasets. Compared to the method proposed in \cite{Lee2014}, the proposed method using individual training datasets shows a superior performance on the Chars74K-15 dataset (68\% versus 64\%) and a comparable performance on the ICDAR03-CH dataset (both achieving a classification accuracy of 79\%). The method presented in \cite{Mishra2013} enriches the training samples by adding extra training samples through affine transforming the original training data. Compared to \cite{Mishra2013}, when using individual training datasets and the linear SVM classifier, the proposed method achieves a much better performance on the ICDAR03-CH dataset (78\% versus 73\%) and a comparable performance on the Chars74K-15 dataset (both achieving a classification accuracy of 68\%). Compared to the method proposed in \cite{Su2014} that uses hybrid training dataset as well, the proposed method achieves comparable performance (both achieving a classification accuracy of 81\%). These results demonstrate the good properties of the HSC features in scene character classification. Using the hybrid training set and the SC classifier, the proposed method achieves a better performance than most competing methods on the ICDAR03-CH dataset (achieving a classification accuracy of 81\%), and achieves the same performance on the Chars74K-15 dataset. We also note that the method (Deep CNN) recently proposed in \cite{Jaderberg2014} achieves pretty high performance and performs the best so far. The high performance of the method (Deep CNN) is mainly due to the advantages of deep learning with a large number of training samples. We emphasize here that the proposed HSC features for scene text recognition are much simpler and easy to implement.

\subsection{Word Recognition} \label{sub-wordrec}
We also evaluate the performance of the proposed method in word recognition using different character classifiers (i.e., FERNS, SVM, and SC) trained with the hybrid training set (as introduced in Section \ref{charClassification}). The word images of the SVT, ICDAR2003, ICDAR2011 and III5K-Word datasets are used for evaluation. For the ICDAR2003 and ICDAR2011 datasets, we use a lexicon created from all the words in the test set (denoted by I03-Full and I11-Full, respectively), and a lexicon consisting of the ground truth word plus 50 random words from the test set (denoted by I03-50 and I11-50, respectively). For the SVT dataset, we use the lexicons containing 50 words provided by \cite{WangK2011}, denoted by SVT-50. Regarding the III5K-Word dataset, the performance of the proposed method on word recognition with 50 words and the Medium lexicon (containing 1000 words for each image) provided by the authors of \cite{Mishra2012a} is evaluated (denoted by III5K-50 and III5K-Med, respectively).

\subsubsection{Word Recognition Performance of the Proposed Method Using Different Features} \label{compFeat}
We first evaluate the word recognition performance of the proposed method using the HSC features and the HOG features. For extracting the HSC features, we set the patch size of the elements in a dictionary to $9\times9$ and set the dictionary size to 100 to better represent character features and get higher performance (see Section \ref{effectsSize} for the influence of the patch size and the dictionary size). The recognition results are shown in Table \ref{wordRecResults}. From the results, we can see that:
\begin{itemize}
\item[(a)] The HSC features outperform the HOG features significantly in word recognition. For all the classifiers, the HSC features outperform the HOG features by a large margin on all the seven experimental settings (i.e., SVT-50, I03-50, I03-Full, I11-50,  I11-Full, III5K-50, and III5K-Med). When using the FERNS classifier, the gained improvement of the HSC features over  the HOG features is about 6\%-8\% on SVT-50, I03-50, and I11-50, and is about 11\%-15\% on the other settings. These results show that the HSC features can more effectively represent character features and structures than the HOG features.
\item[ (b)] When the SC classifier is used, as the number of basis vectors increases, the performance of the proposed method generally gets better. These results indicate that, the SC classifier with a larger number of basis vectors can better reconstruct a character sample and has more discriminative power.
\item[(c)] Using the SC-100 classifier and the HSC features, the proposed method achieves the highest recognition accuracies on I03-50 (92.90\%), I03-Full (87.31\%), I11-50 (93.87\%), I11-Full (89.03\%), III5K-50 (86.70\%), and III5K-Med (73.63\%), which are much higher than those achieved by the state-of-the-art methods (see Section 5.2.3 for more details). On SVT-50, the recognition accuracy obtained by the proposed method using the SC-50 classifier can achieve the highest recognition accuracy (83.31\%).
\end{itemize}

\linespread{1}
\begin{table*} [htpb]
\small
\caption{Word recognition results obtained by the proposed method using the HOG/HSC features and different classifiers (\%). }
\label{wordRecResults}
\begin{center}
\begin{tabular}{|l|l|ccccccc|}
\hline
 &  & SVT-50 & I03-50 & I03-Full  & I11-50 & I11-Full & III5K-50 & III5K-Med \\
\hline
\multirow{2}*{FERNS} & HOG & 64.76 & 78.70 & 65.08  & 80.62 & 66.67 & 64.03 & 46.07 \\
 & HSC & 72.49 & 86.61 & 77.76  & 86.45 & 77.85 & 76.97 & 61.53 \\
\hline
\multirow{2}*{SVM} & HOG & 66.46 & 82.54 & 73.22  & 86.13 & 75.59 & 76.07 & 59.37 \\
 & HSC & 73.88 & 88.36 & 81.84  & 88.92 & 82.04 & 77.79 & 64.85 \\
\hline
\multirow{2}*{SC-10} & HOG & 67.70 & 86.15 & 75.90  & 86.99 & 78.06 & 76.60 & 63.00 \\
 & HSC & 76.20 & 90.10 & 79.05  & 88.82 & 78.39 & 78.73 & 64.40 \\
\hline
\multirow{2}*{SC-20} & HOG & 73.11 & 87.66 & 77.65  & 89.14 & 81.29 & 78.73 & 65.47 \\
 & HSC & 80.22 & 90.45 & 80.91  & 91.08 & 82.26 & 83.80 & 68.97 \\
\hline
\multirow{2}*{SC-30} & HOG & 75.73 & 87.78 & 79.86  & 89.68 & 82.69 & 79.37 & 67.40 \\
 & HSC & 82.07  & 91.39 & 82.77 & 92.04 & 83.76 & 83.50 & 70.07 \\
\hline
\multirow{2}*{SC-50} & HOG & 79.60 & 89.06 & 81.26  & 90.43 & 84.62 & 80.33 & 68.87 \\
 & HSC & \textbf{83.31}  & 91.97 & 84.98 & 93.33 & 86.99 & 85.00 & 72.27 \\
\hline
\multirow{2}*{SC-100} & HOG & 80.53 & 88.71 & 81.37  & 91.51 & 85.16 & 83.00 & 71.10 \\
 & HSC & 83.15 &  \textbf{92.90} &  \textbf{87.31}  &  \textbf{93.87} &  \textbf{89.03} &  \textbf{86.70} &  \textbf{73.63} \\
\hline
\end{tabular}
\end{center}
\end{table*}



\subsubsection{Influence of the Patch Size and the Dictionary Size} \label{effectsSize}

In this section, we evaluate the influence of both the patch size of the elements of the dictionary and the dictionary size on the performance of proposed method in word recognition. In Section \ref{compFeat}, we have shown the performance of the proposed method with the patch size $9\times 9$ and the dictionary size 100 (denoted by HSC\scriptsize{9$\times$9}\normalsize(100)). In this section, we first test the influence of the patch size by evaluating the performance of the proposed method with the patch size being $3\times 3$, $5\times 5$, and $7\times 7$ while fixing the dictionary size to 100 (denoted by HSC$_{3\times3}$(100), HSC$_{5\times5}$(100), and HSC$_{7\times7}$(100), respectively). We then test the influence of the dictionary size by evaluating the performance of the proposed method with the dictionary size being 25, 50, and 75 while fixing the patch size to $9\times 9$ (denoted by HSC$_{9\times9}$(25), HSC$_{9\times9}$(50), and HSC$_{9\times9}$(75), respectively). For all these settings, the classifier used is the SC-100 classifier.

\begin{figure*}[htpb]
\centering
\subfigure[]{
\begin{minipage}[t]{0.47\textwidth}
\includegraphics[width=1\textwidth]{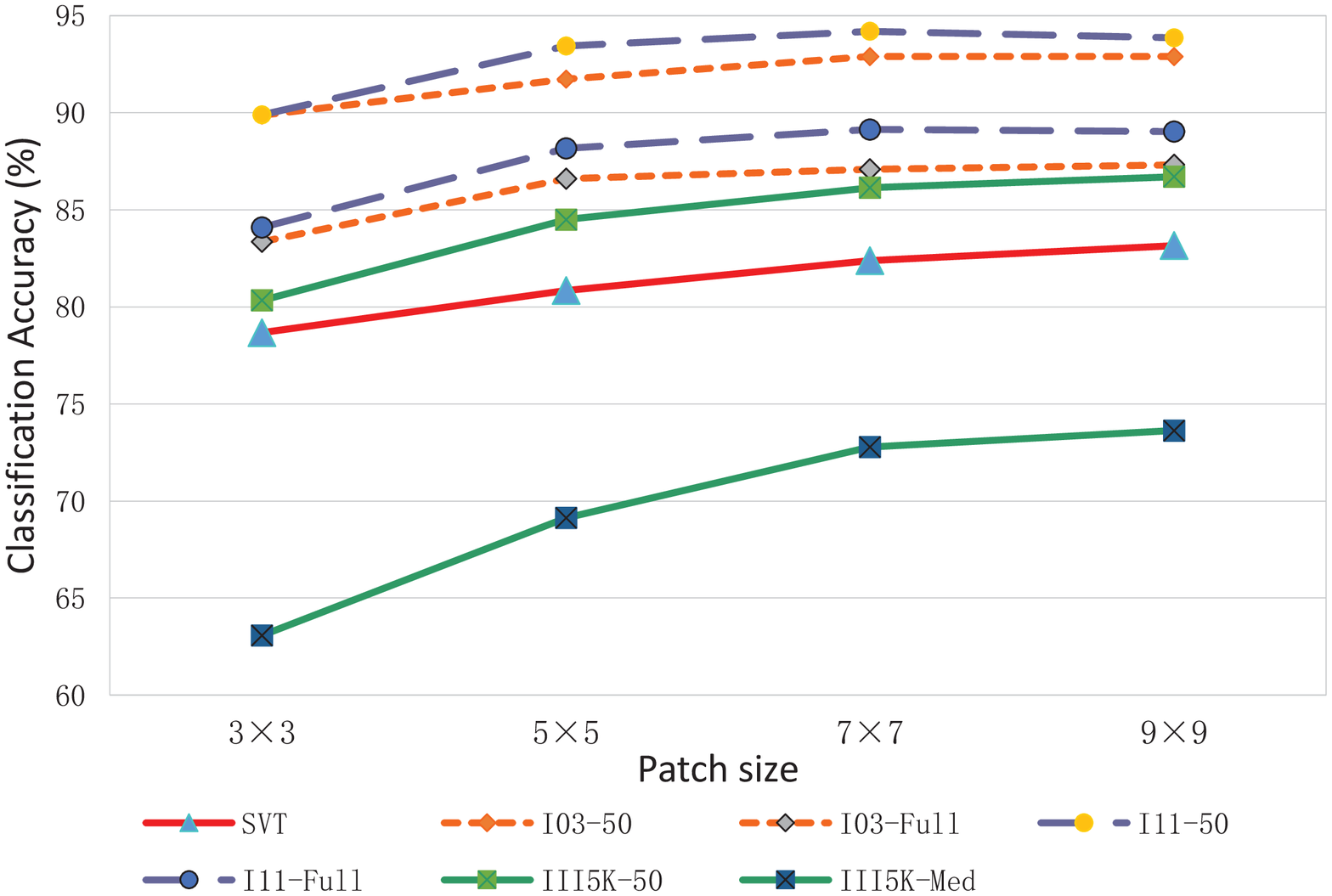}
\label{EffectsPatch.eps}
\end{minipage}
}
\subfigure[]{
\begin{minipage}[t]{0.50\textwidth}
\includegraphics[width=1\textwidth]{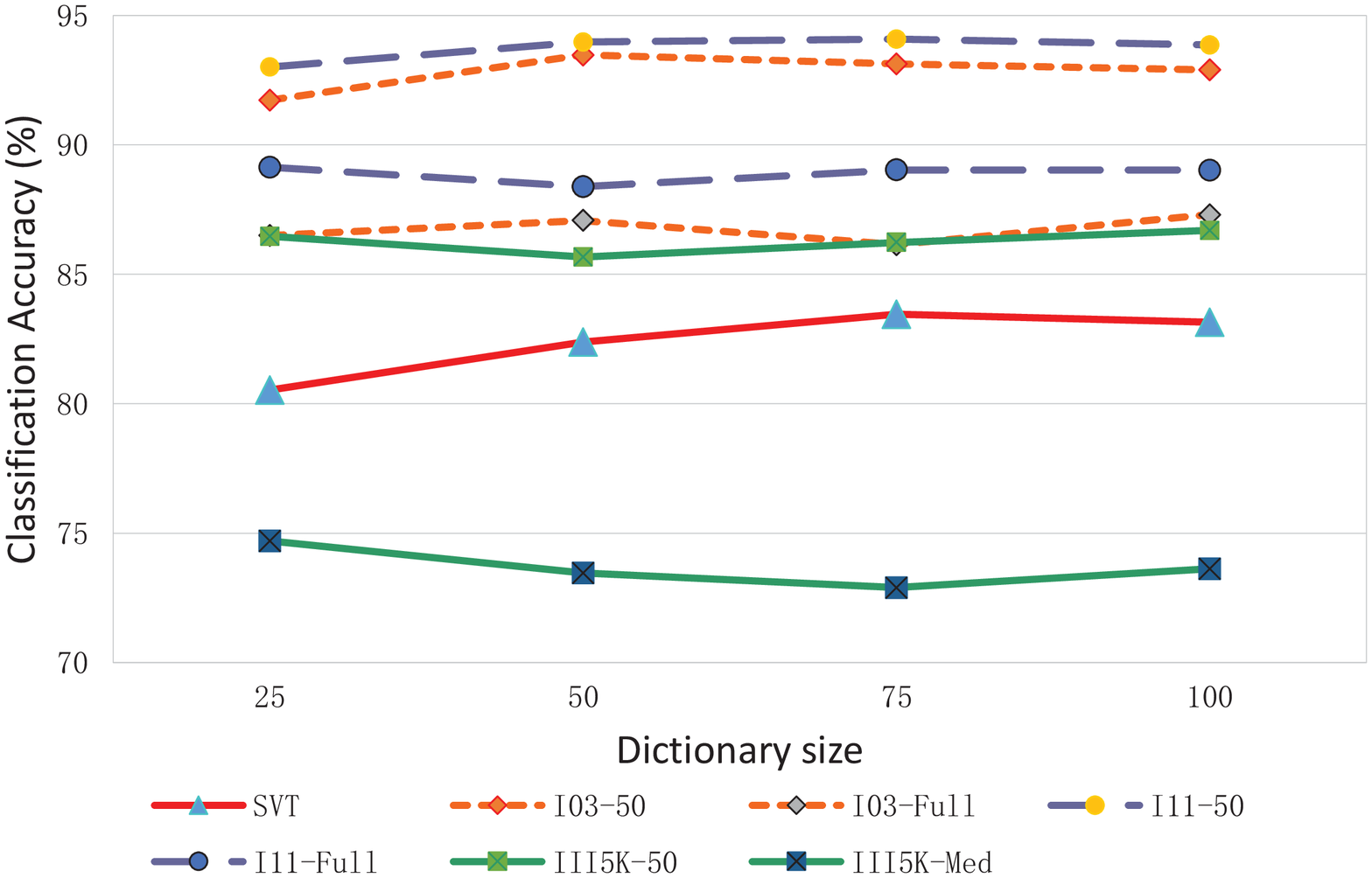}
\label{EffectsDictionary.eps}
\end{minipage}
}
\caption{ Influence of the patch size (a) and the dictionary size (b) on the performance of the proposed method. }
\label{effects}
\end{figure*}

The word recognition results obtained by the proposed method with different patch sizes (while the dictionary size is fixed as 100) are shown in Table \ref{wordRecPat}. From the table, we can see that increasing the patch size from $3\times 3$ to $9\times 9$ causes to increase the recognition accuracy of the proposed method. 
The increase of recognition accuracy from HSC$_{3\times3}$(100) to HSC$_{5\times5}$(100) is quite large, while the increase of that from HSC$_{5\times5}$(100) to HSC$_{7\times7}$(100) is relatively small and  the increase of that from HSC$_{7\times7}$(100) to HSC$_{9\times9}$(100) gets smaller. On I11-50 and I11-Full, the proposed method with  HSC$_{7\times7}$(100) performs even slightly better than that with HSC$_{9\times9}$(100). These results indicate that, though further increasing the patch size may gain slightly better performance, the HSC features extracted using the patch size $9\times 9$ can effectively represent most structural information in characters. Hence in our experiments, we use the patch size $9\times 9$ by default. To better illustrate the influence of the patch size, we show the recognition accuracies of the proposed method with various patch sizes on different datasets in Figure \ref{EffectsPatch.eps}.

\linespread{1}
\begin{table*} [htpb]
\small
\caption{Word recognition results obtained by the proposed method with different patch sizes (\%). }
\label{wordRecPat}
\begin{center}
\begin{tabular}{|l|ccccccc|}
\hline
 & SVT-50 & I03-50 & I03-Full & I11-50 & I11-Full & III5K-50 & III5K-Med \\
\hline
HSC$_{3\times3}$(100) & 78.67 & 89.87 & 83.35 & 89.89 & 84.09 & 80.33 & 63.07 \\
HSC$_{5\times5}$(100) & 80.83 & 91.73 & 86.61 & 93.44 & 88.17 & 84.50 & 69.13 \\
HSC$_{7\times7}$(100) & 82.38 & \textbf{92.90} & 87.08 & \textbf{94.19} & \textbf{89.14} & 86.13 & 72.77 \\
HSC$_{9\times9}$(100) & \textbf{83.15} & \textbf{92.90} & \textbf{87.31} & 93.87 & 89.03 & \textbf{86.70} & \textbf{73.63} \\
\hline
\end{tabular}
\end{center}
\end{table*}

\linespread{1}
\begin{table*} [htpb]
\small
\caption{Word recognition results obtained by the proposed method with different dictionary sizes (\%). }
\label{wordRecDic}
\begin{center}
\begin{tabular}{|l|lllllll|}
\hline
 & SVT-50 & I03-50 & I03-Full & I11-50 & I11-Full & III5K-50 & III5K-Med \\
\hline
HSC$_{9\times9}$(25) & 80.53 & 91.73 & 86.50 & 93.01 & \textbf{89.14} & 86.47 & \textbf{74.70} \\
HSC$_{9\times9}$(50) & 82.38 & \textbf{93.48} & 87.08 & 93.98 & 88.39 & 85.67 & 73.47 \\
 HSC$_{9\times9}$(75) & \textbf{83.46} & 93.13 & 86.15 & \textbf{94.09} & 89.03 & 86.23 & 72.90 \\
HSC$_{9\times9}$(100) & 83.15 & 92.90 & \textbf{87.31} & 93.87 & 89.03 & \textbf{86.70} & 73.63 \\
\hline
\end{tabular}
\end{center}
\end{table*}

Fixing the patch size as $9\times 9$, we evaluate the influence of the dictionary size on the performance of the proposed method in word recognition with different dictionary sizes. Table \ref{wordRecDic} shows the word recognition performance of the proposed method with HSC$_{9\times9}$(25), HSC$_{9\times9}$(50), HSC$_{9\times9}$(75), and HSC$_{9\times9}$(100). From the results, we can see that increasing the dictionary size does not certainly increase the recognition accuracy. However, we also observe that, the performance of the proposed method with the dictionary size being 100 is more stable than that of the proposed method with the dictionary size being the other three sizes (25, 50, 75).  This can be observed from the results that the recognition accuracies of the proposed method using  HSC$_{9\times9}$(100) are only slightly lower than the best results on SVT-50 (HSC$_{9\times9}$(75)), I03-50 (HSC$_{9\times9}$(50)), I11-50 (HSC$_{9\times9}$(75)), I11-Full (HSC\scriptsize{9$\times$9}\normalsize(25)), and III5K-Med (HSC$_{9\times9}$(25)). However, compared with the recognition accuracies obtained by the proposed method with the dictionary size being 100, the recognition accuracies of the proposed method using the other three dictionary sizes (25, 50, and 75) are less stable. For example, although the proposed method using HSC$_{9\times9}$(25) performs the best on I11-Full  and III5K-Med and  comparably with the best on III5K-50 (HSC$_{9\times9}$(100)), the performance of the proposed method using HSC$_{9\times9}$(25) is much lower than the best results achieved on SVT-50 (by HSC$_{9\times9}$(75)), I03-50  (by HSC$_{9\times9}$(50)), I03-Full (by HSC$_{9\times9}$(100)), and I11-50 (by HSC$_{9\times9}$(75)). Thus, we set the dictionary size to 100 by default. 

\subsubsection{Comparing the Proposed Method with Several State-of-the-art Methods} \label{compPre}

We also compare the proposed method with several state-of-the-art methods, and show the results in Table \ref{compState}. We report the performance of the proposed method using the HSC features and the SC-100 classifier as the obtained results by the proposed method. We note that Bissacco et al. \cite{Bissacco2013} and Jaderberg et al. \cite{Jaderberg2014} achieve higher performance than our method and other competing methods. The method proposed in \cite{Jaderberg2014} achieves the highest performance on SVT-50, I03-50, and I03-Full, which is mainly due to the high performance of the CNN based classifier (as indicated in Table \ref{compCharClsf}). However, it is worth noting that, both the two papers use a large number of additional outside training data, while we use all publicly available training data in this paper. Almaz\'{a}n et al. \cite{Almazan2014} achieved highest performance on SVT-50, III5K-50, and III5K-Med (achieving accuracy of 87.01\%, 88.57\% and 75.60\% on SVT-50, III5K-50, and III5K-Med, respectively)using a different framework called word/label embedding (however, they do not report results on the ICDAR2003 and ICDAR2011 datasets). In the following, we mainly compare the proposed method with the other methods. 

\linespread{1}
\begin{table*}[htpb]
\small
\caption{Comparing the proposed method with the state-of-the-art methods on word recognition.}
\label{compState}
\begin{center}
\begin{tabular}{|c|c|c|c|c|c|c|c|c|}
  \hline
   Method & SVT-50  & I03-50 & I03-Full  & I11-50 & I11-Full  & III5K-50 & III5K-Med  \\  \hline
   K. Wang et al. \cite{WangK2011} & 57 & 76 & 62 & -- & -- & -- & -- \\
    Mishra et al. \cite{Mishra2012a} & 73.26 & 81.78 & -- & -- & -- & 68.25  & 55.50 \\
    Mishra et al. \cite{Mishra2012b} &73.57& 80.28 & -- &-- & -- & 66 &57.5 \\
    Novikova et al. \cite{Novikova2012} & 72.9 & 82.8 & --  & -- & -- & -- & -- \\
    T. Wang et al. \cite{WangT2012} & 70 & 90 & 84 & -- & -- & -- & -- \\
    Shi et al. \cite{Shi2013} & 73.51 & 87.44 & 79.30 & 87.04 & 82.87 & -- & -- \\
    Goel et al. \cite{Goel2013} & 77.28 & 89.69 & -- & -- & -- & -- & -- \\
    Weinmann et al. \cite{Weinman2014} & 78.05 & -- & -- & -- & -- & -- & -- \\
    Shi et al. \cite{Shi2014a} &74.65 & 84.52 & 79.98 & -- & -- & -- & -- \\
    Shi et al. \cite{Shi2014b} & 73.67 & 87.83 & 79.58 & 87.22 & 83.21 & -- & -- \\
    Yao et al. \cite{Yao2014} & 75.89 & 88.48 & 80.33  & -- & -- &80.2 &69.3 \\
    Lee et al. \cite{Lee2014} &80 &88 &76 &88 &77 &-- &-- \\
    Su et al. \cite{Su2014b}   & 83 & 92 & 82 & 91 & 83 &  -- &  -- \\
    Almaz\'{a}n et al. \cite{Almazan2014}  & \textbf{87.01} & -- &  -- & -- &  -- & \textbf{88.57} & \textbf{75.60} \\
    Jaderberg et al. \cite{Jaderberg2014} & 86.1 & \textbf{96.2} &\textbf{91.5} & -- & --& --& --\\ \hline
  \textbf{Proposed} HSC+SC  & \textbf{83.15} &\textbf{92.90}	&\textbf{87.31}	&\textbf{93.87}	&\textbf{89.03}	&\textbf{86.70}	&\textbf{73.63}  \\ \hline
\end{tabular}
\end{center}
\end{table*}

From Table \ref{compState}, we can see that the proposed method outperforms most the competing state-of-the-art methods on all the datasets. On SVT-50, the recognition accuracy obtained by the proposed method is higher than that achieved by \cite{Lee2014} (83.15\% versus 80\%). On I03-50 and I03-Full, the proposed method achieves 3\% higher recognition accuracy when it is compared to \cite{WangT2012} which uses CNN as the character classifier. On I11-50 and I11-Full, the proposed method obtains an improvement of 6\%-7\% higher recognition accuracy when compared to \cite{Shi2013} and \cite{Shi2014b}. When compared to \cite{Lee2014} on I11-50 and I11-Full, the proposed method obtains an improvement of 6\% higher recognition accuracy on I11-50 (93.87\% versus 88\%) and an improvement of 12\% higher recognition accuracy on I11-Full (89\% versus 77\%). On III5K-50 and III5K-Med, the recognition accuracies obtained by the proposed method  are much higher than those obtained by \cite{Yao2014} (86.70\% versus 80.2\% on III5K-50 and 73.63\% versus 69.3\% on III5K-Med). When compared to \cite{Su2014b}, the proposed method performs comparably on SVT-50 and I03-50, and achives higher performance on I03-Full, I11-50, and I11-Full (87.31\% versus 82\% on I03-Full, 93.87\% versus 91\% on I11-50, and 89.03\% versus 83\% on I11-Full). These results demonstrate the effectiveness of the proposed HSC-based scene text recognition method.

\subsection{Recognition Speed}

In this section, we will evaluate the recognition speed of  the proposed method in character classification and word recognition, and report the average CPU time for per character/word image sample. For character classification, we test the process time of  the proposed method in feature extraction and feature classification  separately. For word recognition, we evaluate the process time in the character detection step and the DP search step separately.

\subsubsection{The Computational Speed of the Proposed Method in Character Classification}

We first compare the process time of the proposed method in extracting the HSC features and the HOG features. For the HSC features, we evaluate the HSC features with different dictionary sizes and a fixed patch size of $9\times 9$ (i.e., HSC$_{9\times9}$(25), HSC$_{9\times9}$(50), HSC$_{9\times9}$(75), and HSC$_{9\times9}$(100)), and different patch sizes and a fixed dictionary size of 100 (i.e., HSC$_{3\times3}$(100), HSC$_{5\times5}$(100), HSC$_{7\times7}$(100), and \\HSC$_{9\times9}$(100)). The CPU time used in extracting the HOG features is 1.3 millisecond for each character sample. Table \ref{speedFE} shows the CPU time used in extracting the HSC features (with different dictionary sizes and patch sizes) for each character sample. From the results, we can see that the feature extraction step of the HSC features is more computationally expensive than that of the HOG features. For the HSC features, as the patch size or the dictionary size grows, the computation cost used in extracting the HSC features increases.

\linespread{1}
\begin{table*}[htpb]
\small
\caption{The CPU time used in the HSC feature extraction per character sample (in millisecond).}
\label{speedFE}
\begin{center}
\begin{tabular}{|l|cccc|}
\hline
 HSC with different dictionary sizes  & HSC$_{9\times9}$(25) & HSC$_{9\times9}$(50)  & HSC$_{9\times9}$(75) &HSC$_{9\times9}$(100) \\ \hline
  Time/sample (ms)  & 3.3 & 4.2 & 5.2 & 6.0  \\		\hline \hline
HSC with different patch sizes  & HSC$_{3\times3}$(100) &  HSC$_{5\times5}$(100) & HSC$_{7\times7}$(100) & HSC$_{9\times9}$(100) \\ \hline
 Time/sample (ms) &2.9    &3.6    & 4.5 &	6.0	\\
\hline
\end{tabular}
\end{center}
\end{table*}

\begin{figure}[htbp]
\centering
\includegraphics[width=0.450\textwidth]{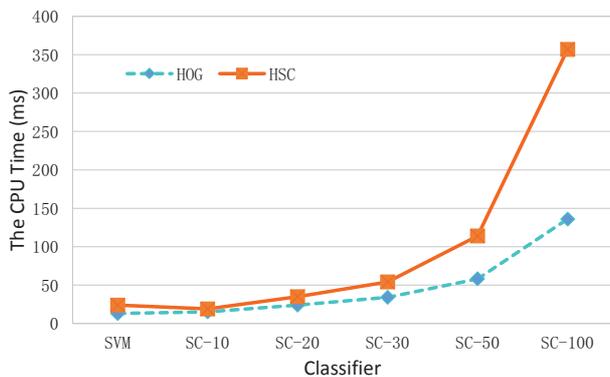}
\caption{  The CPU time of the proposed method in feature classification per character sample.  }
\label{SpeedCharClsf}

\label{examples}
\end{figure}

We then evaluate the CPU time of the proposed method used in classifying the extracted features. For the HSC features, we set the dictionary size to 100 and the HSC features fed to the trained classifier have a dimensionality of $1600$ (see Section \ref{sec-clsfTrn} for more details). The dimensionality of the HOG features is $1116$ (see Section \ref{sec-clsfTrn} for more details). The feature classification speed of the proposed method using different features and classifiers is shown in Table \ref{speedClsf} (as well as in Figure \ref{SpeedCharClsf} for better illustration). We can see that the FERNS classifier is much faster (about 100-1000 times faster) than the other classifiers (of course, the proposed method using the FERNS classifier achieves much lower accuracy than that with the other classifiers. See Table \ref{charClsf} and \ref{charClsf2} for more details). For the SC classifier, increasing the number of the basis vectors of the classifier increases the computation cost. Table \ref{speedFE} and \ref{speedClsf} also show that, the feature classification step is much more computationally expensive than the feature extraction step.

\linespread{1}
\begin{table*}[htpb]
\small
\caption{The CPU time of the proposed method used in feature classification per character sample using HOG/HSC features and different classifiers (in millisecond).}
\label{speedClsf}
\begin{center}
\begin{tabular}{|l|l|lllllll|}
\hline
 &  & FERNS & SVM & SC-10 & SC-20 & SC-30 & SC-50 & SC-100 \\
\hline
Time/sample (ms) & HOG & 0.08 & 13 & 15 & 24 & 34 & 58 & 136 \\
 & HSC & 0.1 & 24 & 19 & 35 & 54 & 114 & 357 \\
\hline
\end{tabular}
\end{center}
\end{table*}

\subsubsection{The Computational Speed of the Proposed Method in Word Recognition}

\begin{figure}[htbp]
\begin{center}
  \includegraphics[width=0.45\textwidth]{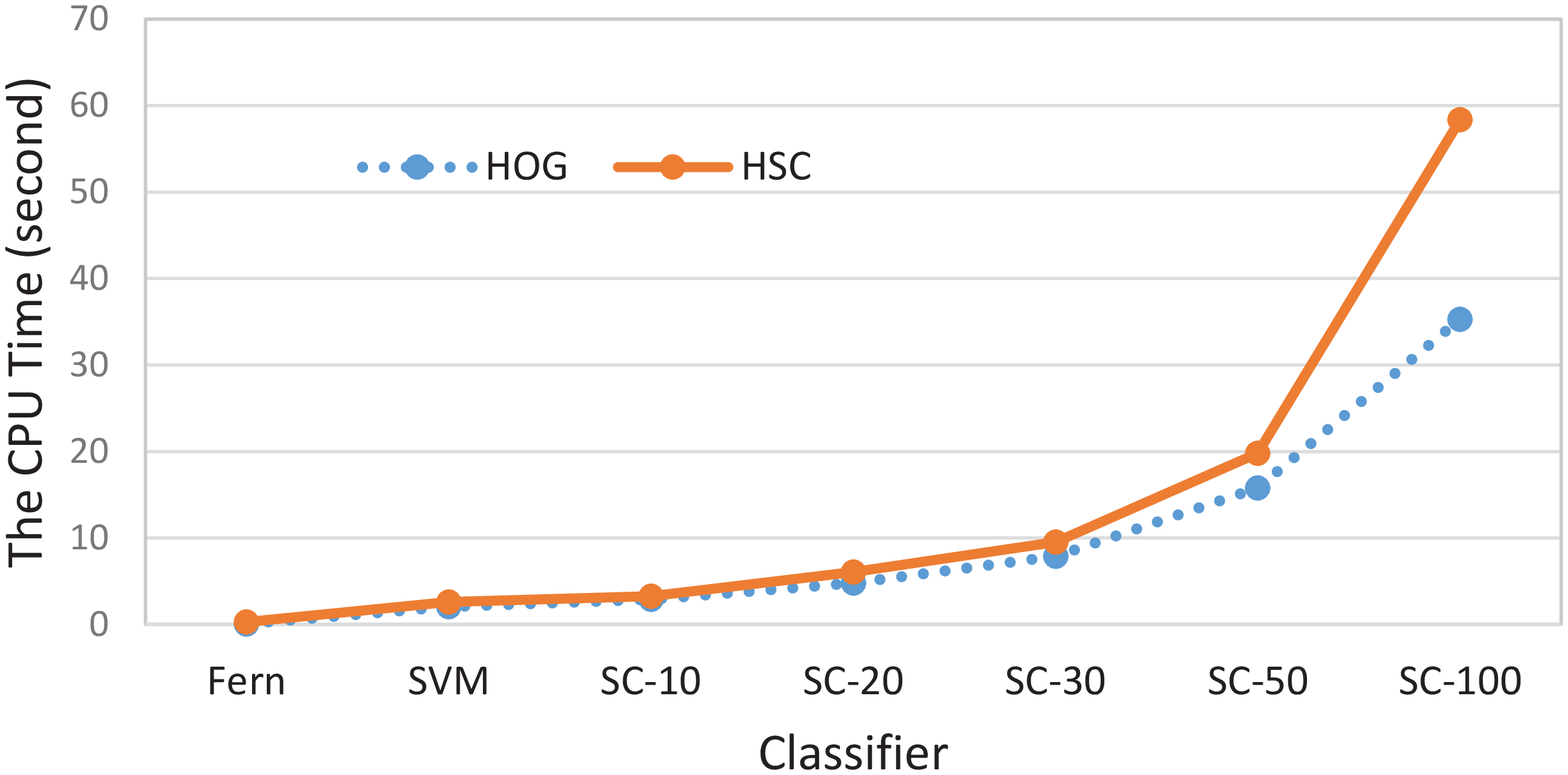}
  \caption{The CPU time used in the character detection step of the proposed method using different features and classifiers (per word sample).}
  \label{SpeedCD.eps}
\end{center}
\end{figure}

In this section, we test the CPU time of the proposed method in word recognition. We report the average CPU time used in the two steps (i.e., character detection and the DP search) separately. The computation cost of the proposed method used in the character detection step depends on the used feature extraction method and character classifier. For the HSC features, we fix the patch size to $9 \times 9$ and the dictionary size to $100$. The CPU time used in the character detection step of the proposed method using different classifiers is shown in Table \ref{speedCharDet} (as well as in Figure \ref{SpeedCD.eps} for better illustration). From the result, we can see that the proposed method using the FERNS classifier performs much faster than that using the other classifiers. When the SC classifier with the number of the basis vectors being more than 20 is used, the character detection step of the proposed method is slow, which is mainly due to the high computation cost in classifying a large number of character candidates. 

\linespread{1}
\begin{table*}[htpb]
\small
\caption{The CPU time used in the character detection step of the proposed method using different features and classifiers (per word image sample, in second).}
\label{speedCharDet}
\begin{center}
\begin{tabular}{|l|lllllll|}
\hline
 & FERNS & SVM & SC-10 & SC-20 & SC-30 & SC-50 & SC-100 \\
\hline
HOG & 0.078 & 2.06 & 2.91 & 4.79 & 7.87 & 15.81 & 35.31 \\
HSC  & 0.285 & 2.59 & 3.24 & 6.05 & 9.55 & 19.82 & 58.35  \\
\hline
\end{tabular}
\end{center}
\end{table*}

The CPU time used in the DP search step of the proposed method mainly depends on the number of words contained in the lexicon. Since the numbers of words in the lexicons of I03-Full, I11-Full, and III5K-Med are nearly the same (about 1000 words), we report the CPU time of the proposed method using lexicons with 50 words and the Full lexicons (without discriminating between different datasets). In the experiments, the average CPU time used in the DP search step of the proposed method per word image is 1.05 second for 50 words and 8.68 second for the Full lexicons, respectively. Since the proposed mthod adopts the word spotting strategy for word recognition, using the Full lexicons is much slower than using only 50 words in word recognition, which shows that using the word spotting strategy for word recognition is only suitable for small/medium lexicons.

These results show that, for word recognition, the recognition speed of the proposed method using the multi-scale sliding window strategy is still slow when using more powerful feature extraction and classification methods. One solution to this problem is to apply a general object detector to obtain preliminary character candidates and then recognize these character candidates using more effective feature extraction and classification algorithms. The general object detector needs to find as more character candidates as possible using only a small number of candidate proposals. An alternative solution is to use dimensionality reduction techniques, such as principal component analysis (PCA), to reduce the feature dimensionality so as to reduce the computation cost in character detection. This will benefit a lot to the SC classifier with a larger size of basis vectors. In this paper, we mainly focus on using sparse coding based features for scene text recognition, thus we do not evaluate the influence of dimensionality reduction techniques on the performance of the proposed method in word recognition.

\subsection{Text Recognition Examples}

\begin{figure*}[!htb]
  \centering
  \includegraphics[width=1\textwidth]{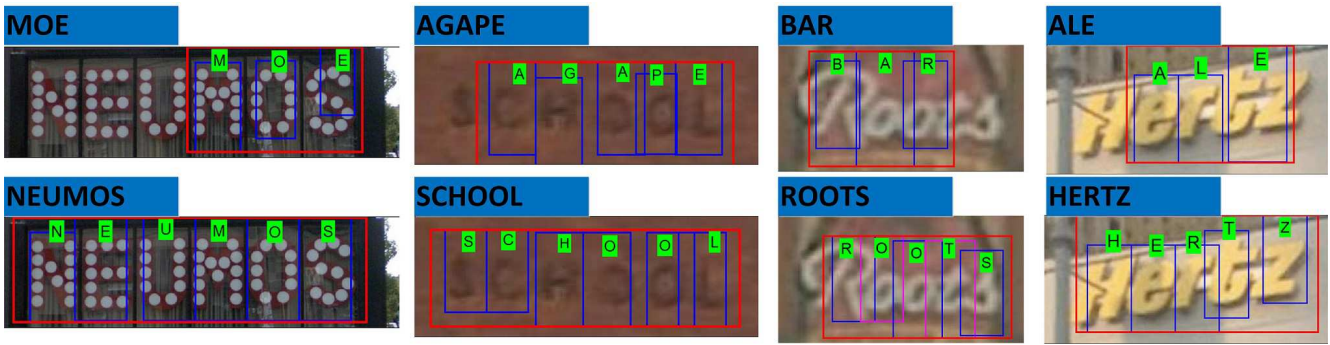}
  \caption{Some recognition results obtained by the proposed method on the SVT dataset. The upper row shows the results of the proposed method using the HOG features, and the lower row shows those using the HSC features. }
  \label{RecognitionExamplesSVT.eps}
\end{figure*}

\begin{figure*}[!htb]
  \centering
  \includegraphics[width=1\textwidth]{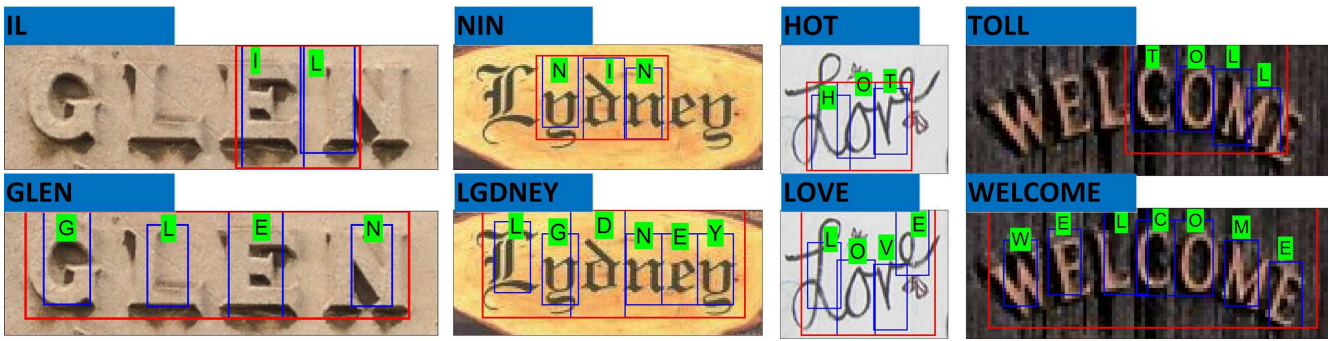}
  \caption{Some recognition results obtained by the proposed method on the III5K-Word dataset. The upper row shows the results of the proposed method using the HOG features, and the lower row shows those using the HSC features. }
  \label{RecognitionExamples5k.eps}
\end{figure*}

In this section, we show and analyze some recognition results obtained by the proposed method. We first provide some recognition results of the proposed method with the HOG features and the HSC features using the SC-100 classifier. Some recognition results on the SVT and III5K-Word datesets are shown in Figure \ref{RecognitionExamplesSVT.eps} and Figure \ref{RecognitionExamples5k.eps}, respectively. In the two figures, the upper row shows the results obtained by the proposed method using the HOG features and the lower row shows those using the HSC features. In both figures, the word images are not correctly recognized by the proposed method using the HOG features, but they are successfully recognized by the proposed method using the HSC features.

As shown in Figure \ref{RecognitionExamplesSVT.eps}, the images from the SVT dataset are difficult to recognize due to cluttered background (such as the word ``NEUMOS'' and ``HERTZ''), similar color in background (such as the word ``SCHOOL''), low resolution, shadow (such as the word ``HERTZ''), etc. For some images such as  the word ``SCHOOL'', it is difficult to discriminate the characters from the background even for human beings, but the proposed method using the HSC features can detect and recognize the characters correctly, showing the robustness of the proposed HSC-based scene text recognition method.

In Figure \ref{RecognitionExamples5k.eps}, it shows that the images from the III5K-Word dataset are difficult to recognize due to similar color to the background (such as the word ``GLEN''), the ambiguity caused by irregular characters (e.g., the art characters in the words ``LGDNEY'' and ``LOVE''), rotated characters in curved text lines (such as the word ``WELCOME''), etc. For these words, the proposed method  fails to detect the characters and recognize the words when using the HOG features, while surprisingly, it is able to detect the characters and recognize the words correctly when using the HSC features. For the word ``GLEN'', we can see that although the recognition result using the HOG features is incorrect, it is meaningful in that, the recognition result ``IL'' is reasonable if one views the shadow of the characters as the characters ``I'' and ``L''. In fact, in the III5K-Word dataset, there are many word images with art characters, rotated characters and curved text lines, which makes the dataset challenging. The HSC features outperform the HOG features significantly for word recognition as they can describe much richer information of characters and thus effectively handle the challenges mentioned above.

\begin{figure*}[!htb]
  \centering
  \includegraphics[width=1\textwidth]{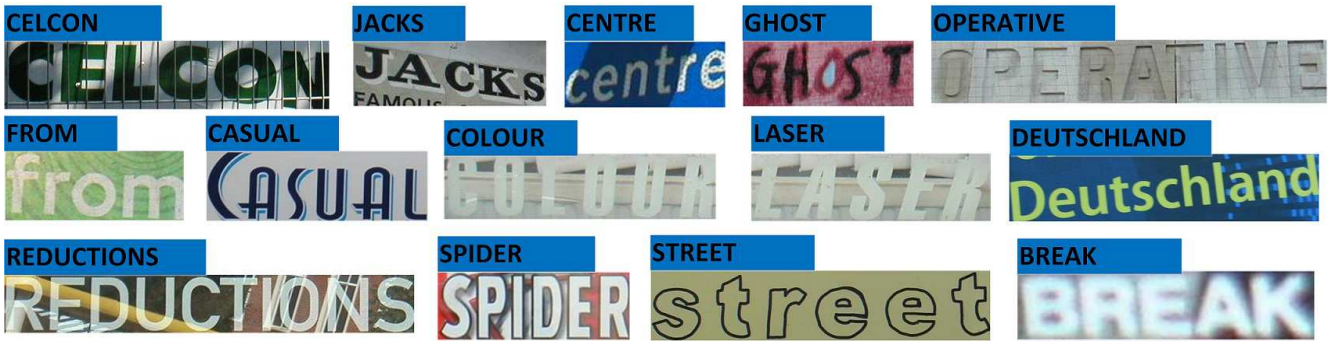}
  \caption{More recognition results obtained by the proposed method on the ICDAR2003 and ICDAR2011 datasets.}
  \label{RecognitionExamplesSuc.eps}
\end{figure*}

\begin{figure*}[!htb]
  \centering
  \includegraphics[width=0.9\textwidth]{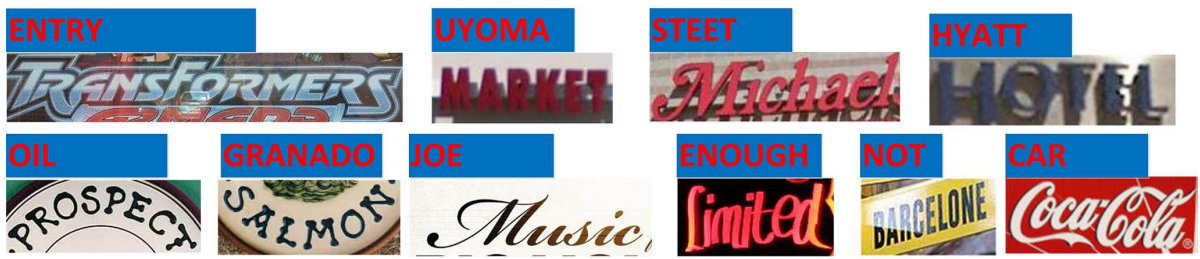}
  \caption{Some recognition examples that are not correctly recognized by the proposed method. The examples in the upper row are from the SVT dataset, and the ones in the lower row are from the III5K-Word dataset.}
  \label{RecognitionExamplesFail.eps}
\end{figure*}

We also show more word recognition results that are correctly recognized by the proposed method on the ICDAR2003 and ICDAR2011 datasets in Figure \ref{RecognitionExamplesSuc.eps}, and show some recognition results that are not correctly recognized by the proposed method on the SVT and III5K-Word datasets in Figure \ref{RecognitionExamplesFail.eps}. Figure \ref{RecognitionExamplesSuc.eps} shows that the proposed HSC-based scene text recognition method can robustly and effectively handle the challenges. 
Figure \ref{RecognitionExamplesFail.eps} shows the limitations of the proposed method that need more efforts to solve, such as to deal with seriously rotated characters in curved text lines, heavily distorted art characters, very low discrimination between characters and the background, etc. Since in this paper we use the multi-scale sliding windows strategy to generate character candidates, the current method is suitable for recognizing near horizontal/vertical text and do not performs well on curved text lines that contain rotated characters. Hence exploiting specific methods for recognizing text lines with various directions will be our future work.

\section{Conclusion} \label{conclusion}

In this paper, we have proposed an effective scene text recognition method using sparse coding based features (i.e., the HSC features), which are extracted by computing per-pixel sparse codes of characters and aggregating the sparse codes to form local histograms. The HSC features can represent much richer structural information of characters, and thus the proposed method using the HSC features significantly outperform the competing methods using the other features (such as HOG, LHOG, GHOG, LBP, GB, etc.) in scene character/text recognition. For word recognition, we propose to integrate character detection results with geometric constraints and use the MCE training method to learn the parameters of the proposed method. Experimental results on several popular datasets (ICDAR2003, ICDAR2011, SVT, and III5K-Word) show the effectiveness and robustness of the proposed method, which outperforms most state-of-the-art methods.

In our future work, two research directions will be considered: One is to use a general object detector to reduce the number of character candidates that need to be recognized by the character classifier; The other one is to use dimensionality reduction techniques to speed up the character detection process. Besides, we will also extend the current work (word recognition) to full image word detection and recognition.

\section*{Acknowledgment}
This work was supported by the National Natural Science Foundation of China under Grants 61305004, 61472334 and 61170179, by China Postdoctoral Science Foundation under Grant 2012M521277.



\end{document}